\documentclass[10pt,twocolumn,letterpaper]{article}

\usepackage{cvpr}
\usepackage{times}
\usepackage{epsfig}
\usepackage{graphicx}
\usepackage{amssymb}
\usepackage{amsmath}

\usepackage[pagebackref=true,breaklinks=true,letterpaper=true,colorlinks,bookmarks=false]{hyperref}

\cvprfinalcopy 

\def\cvprPaperID{****} 

\ifcvprfinal\fi
\begin{document}

\title{Reducing Overlearning through Disentangled Representations by Suppressing Unknown Tasks}

\author{Naveen Panwar, Tarun Tater, Anush Sankaran, Senthil Mani\\
IBM Research AI, India
}

\maketitle

\begin{abstract}
Existing deep learning approaches for learning visual features tend to overlearn and extract more information than what is required for the task at hand. From a privacy preservation perspective, the input visual information is not protected from the model; enabling the model to become more intelligent than it is trained to be. Current approaches for suppressing additional task learning assume the presence of ground truth labels for the tasks to be suppressed during training time. In this research, we propose a three-fold novel contribution: (i) a model-agnostic solution for reducing model overlearning by suppressing all the unknown tasks, (ii) a novel metric to measure the trust score of a trained deep learning model, and (iii) a simulated benchmark dataset, PreserveTask, having five different fundamental image classification tasks to study the generalization nature of models. In the first set of experiments, we learn disentangled representations and suppress overlearning of five popular deep learning models: VGG16, VGG19, Inception-v1, MobileNet, and DenseNet on PreserverTask dataset. Additionally, we show results of our framework on color-MNIST dataset and practical applications of face attribute preservation in Diversity in Faces (DiF) and IMDB-Wiki dataset.
\end{abstract}

\section{Introduction}
With the advent of deep learning (DL), the models are striving to perform composite tasks by learning complex relationships and patterns available in noisy, unstructured data~\cite{ruder2017overview}. Feature entanglement~\cite{lu2019unsupervised, Jiang_2019_CVPR, lee2018diverse} is an observed property, where the features learnt for a specific objective is shown to carry information and properties of other objectives. This is primarily attributed to the learning capacity of the deep learning models and are used effectively in multiple applications for general intelligence such as multi-task learning~\cite{Li_2019_CVPR} and transfer learning~\cite{Tang_2019_CVPR}.
 
\begin{figure*}[!t]
	\begin{center}
		\includegraphics[width=6.4in]{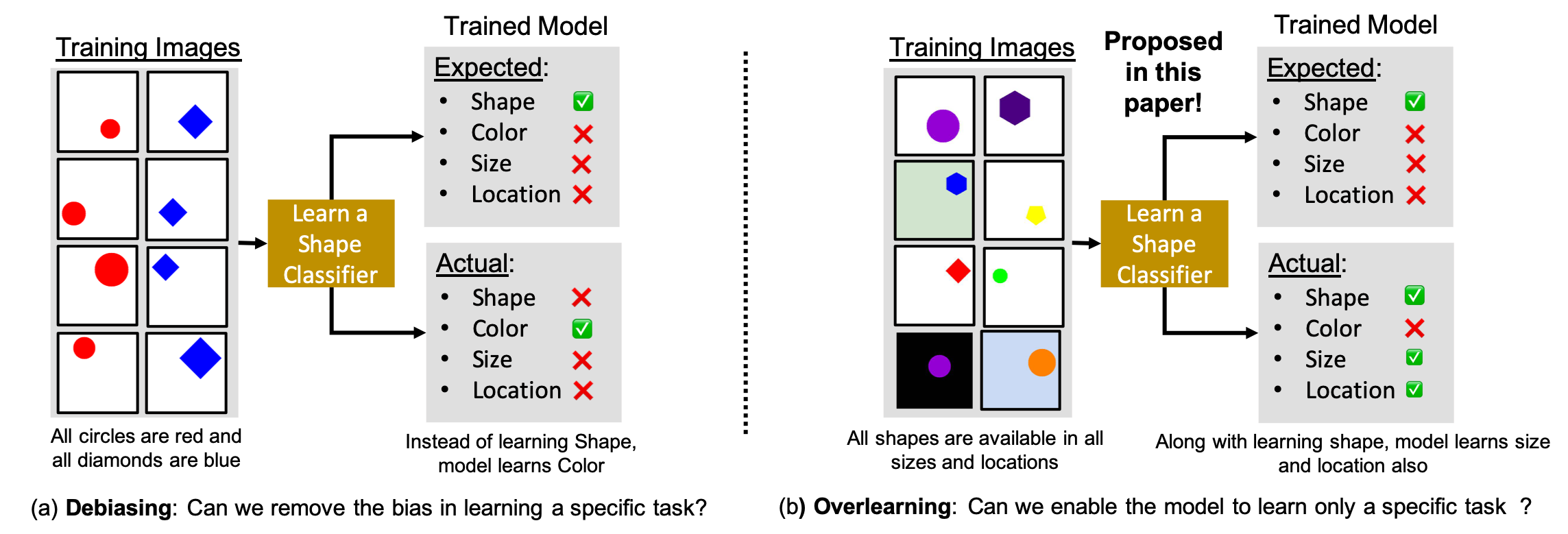}
	\end{center}
	\caption{Visually distinguishing the concepts model debiasing and reducing model overlearning (what we aim to do). The fundamental research motivation in this work is to study if a learning model could be restricted to perform only one objective from a given training dataset.}
	\label{fig:intro1}
\end{figure*}

However from a privacy preserving perspective, the model itself could learn all the private information from the data and become much more intelligent than the original intent it was trained for. This phenomenon is called as model overlearning~\cite{song2019overlearning}. 
Consider the example described in Figure~\ref{fig:intro1} (b), where a DL classifier is trained to detect the shape of an object from images. However using the extracted features, the size and location of the object in the image can also be predicted with sufficient accuracy. Thus, a DL classifier trained only for shape prediction is more intelligent than its objective of only predicting the shape of the object. The features used for predicting the shape and size of the object are highly disentangled, as they share a lot of common properties. Thus, to ensure that a trained DL model performs sufficiently only one task, it is required to disentangle these shared representations by explicit supervision.
As an additional real-world example, we train a DL model to predict the gender from a face image. However, the DL model learns most generic features from the face image, enabling it to predict the age and the identity of the person. In applications where the identity of the person has to be preserved from face image, it is needed that the DL model is trained to suppress the task of identity prediction while still performing gender prediction.

A common fictional example quoted from the movie world is that of \textit{Skynet}~\footnote{(Spoiler Alerts!) The future AI system in the Terminator Franchise. Skynet is currently replaced by Legion in Terminator Dark Fate.} which is a neural network based super AI that becomes self aware and tries to take over the human race. The motivation derived from that work of fiction is that, if AI systems are allowed to be more intelligent and overlearn than what is required for a particular task, it could result in AI models having a better understanding and functioning of the data than humans do.

Additionally as shown in Figure~\ref{fig:intro1} (a), the task of debiasing is to remove the the bias (color) in learning a specific task (shape). This happens due to the high correlation between the color and shapes in the input images. However, as shown in Figure~\ref{fig:intro1} (b), our task in model trust is to forcefully ensure that the model learns to perform only one or few selected tasks (shape) from the input images and unlearn all other tasks (color, size, location).

\subsection{Research Contributions}
If multi-class classification tasks could be done from the same image, the research question is, \textit{``How can we ensure that the model is learnt only for one task (called as, preserved tasks), and is strictly not learnt for the other tasks (called as, suppressed tasks)?"}. 
To pursue research on this problem, there are few evident challenges: (i) there is a lack of a balanced and properly curated image dataset where multiple classification tasks could be performed on the same image, (ii) the complete knowledge of both the preserved tasks and the suppressed tasks should be known \textit{apriori}, that is, we cannot suppress those tasks that we don't have information about, and (iii) presence of very few model agnostic studies to preserve and suppress different task groups. The major research contributions are summarized as follows:
\begin{enumerate}
    \item A generic model-agnostic solution framework to reduce model overlearning by suppressing other tasks with shared entangled features. Feature disentanglement is performed using random unknown classes, breaking the assumption of requiring the ground truth labels for suppression tasks during training.

    \item A metric to measure the trust score of a trained DL model. The trust scores specify the amount of overlearning by the model for other tasks, with higher trust scores denoting suppression of overlearning.

	\item A simulated, class-balanced, multi-task dataset, \textit{PreserveTask} with five tasks that could be performed on each image: shape, size, color, location, and background color classification.
	
	\item Experimental analysis are performed for the proposed framework in comparison with other existing approaches under different settings. We demonstrate the overlearning ability of five different deep learning models: VGG16,VGG19,  Inception-v1, MobileNet, and DenseNet. Also, we show the effectiveness of feature disentanglement by suppressing using unknonwn random tasks.~\footnote{The benchmark dataset along with the splits, baselines features, results, and the code are made available here: \url{https://github.com/dl-model-recommend/model-trust}}.
	
	\item To demonstrate the practical applications and generalizability of the metric and the solution framework, we show additionally results in colored MNIST dataset and face attribute preservation using two datasets: (i) Diversity in Faces (DiF)~\cite{dif_arxiv19} (ii) IMDB-Wiki~\cite{rothe2018deep}.
\end{enumerate}

\section{Literature Review}
There are broadly two different groups of work related to the research problem at hand: (i) k-anonymity preservation and (ii) attribute suppression.

\begin{figure*}[!t]
	\begin{center}
		\includegraphics[width=6.4in]{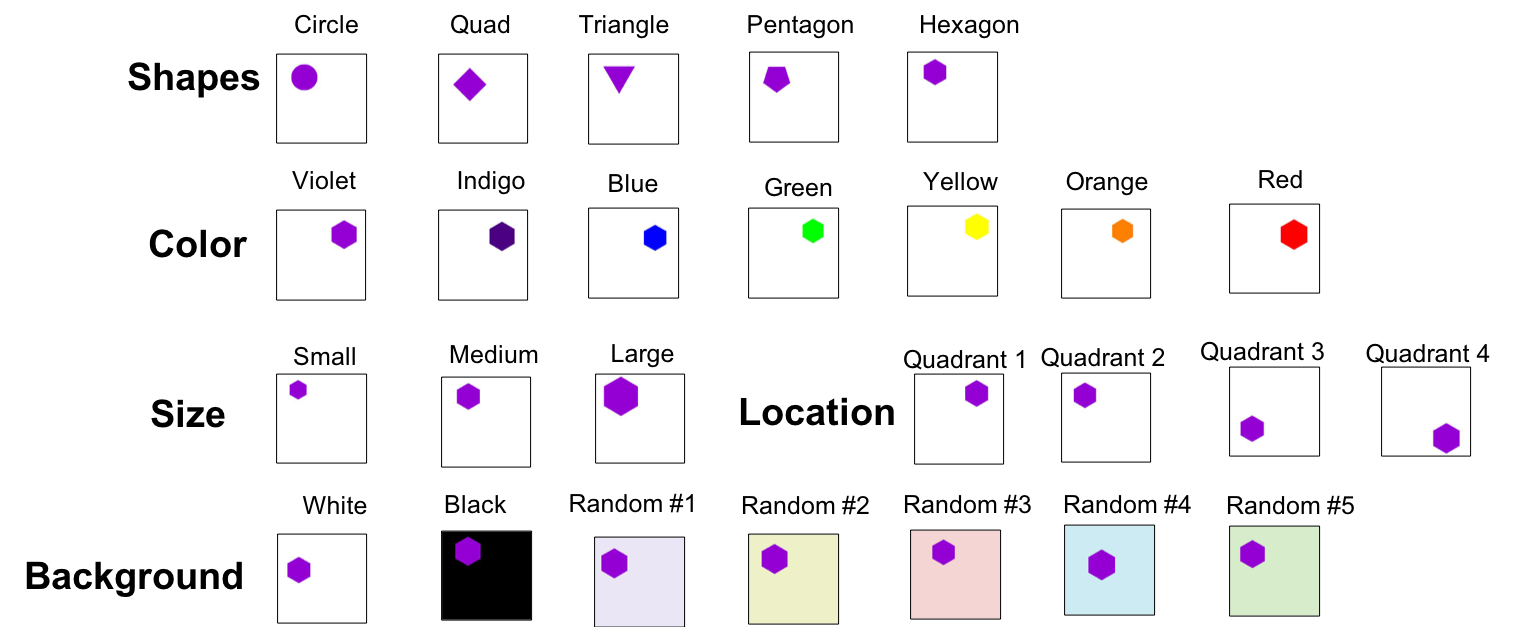}
	\end{center}
	\caption{Landscape of the \textit{PreserveTask} dataset describing the set of different possible tasks. Five tasks could be performed on each image and each task has varying number of classes.}
	\label{fig:dataset}
\end{figure*}

\noindent \textbf{k-anonymity Preservation:} The objective here is to preserve the anonymity of certain attributes from being predicted by the model. To quote some earlier works, ~\cite{boyle2000effects}, studied to mask out potentially sensitive information from video feeds. In the last decade, face recognition has become an important commercial applications and also an application that demanded discussion regarding privacy preservation. Studies focused on extracting only the required meta information from face images while not extracting the identity. This was a required step to make face as a usable biometric. Studies such as \cite{gross2006model}, \cite{newton2005preserving}, and \cite{mirjalili2018semi} focused on preserving the identity of the face image from the model by performing face de-identification. Studies such as \cite{mirjalili2017soft} and \cite{othman2014privacy} focused on anonymizing the face gender information while models could extract the identity.

\noindent \textbf{Attribute Suppression:} The aim of this group of techniques is to explicitly suppress a few attributes by perturbing the input data to the model. Studies such as \cite{rozsa2016facial} and \cite{rozsa2017facial} test if the learnt models are robust and protected against adversarial attacks. \cite{chhabra2018anonymizing} suggested using a constrained generative adversarial network (GAN) to perturb the input face image and suppress the required attribute. The GANs will generate the attribute free face image of the original face image. The closest related work to our approach, is the study by \cite{jayaraman2014decorrelating} where the visual attributes are decorrelated using a negative gradient in the model. The results demonstrate that the classification task could be performed by preserving specific attributes in the image while suppressing the influence of the remaining.

Additionally, there is a good amount of research in bias mitigation while learning models~\cite{zhao2017men}~\cite{kim2018learning}~\cite{attenberg2015beat}~\cite{li2019repair}. The primary aim is to debias the model learning from any kind of correlated attributes~\cite{alvi2018turning}~\cite{raff2018gradient}~\cite{kim2019learning}~\cite{wang2019privacy}, which is different from our aim of improving the model's trust.
The major gaps in the existing research works are: (i) most of the techniques focus on data perturbation, that is, changing the input data from $x$ to $x'$ such that the suppressed task information is not available in the data. There is not much focus on model perturbation without altering the input data, (ii) most of the existing datasets have only binary attributes and hence suppressing and preserving a few tasks does not actually translate to the classification complexity of multi-class tasks, and (iii) there is a lack of a well curated benchmark dataset to evaluate the privacy preserving capacity of DL models. 
\section{PreserveTask Dataset}

Shared tasks performed on the same image carry some common attributes which are often extracted by complex deep learning models. 
The objective of this is to untangle the shared tasks and enable deep learning models to perform only one (or few) of those tasks. In order to evaluate the performance of such a framework, the dataset should have the following properties:
\begin{itemize}
	\item Should perform multiple tasks on the same image and each task should have varying number of classes, in order to study the relationship of complexity of classification tasks.
	\item As this research area is nascent, the dataset should be noise-free and class balanced, to avoid other complexities that could influence classification performance. 
	\item Tasks should be designed in such a way that certain tasks, share common attributes and features, while certain tasks should be independent of each other.
\end{itemize}

There are some similar publicly available datasets in the literature. LFW~\cite{learned2016labeled}, CelebA~\cite{liu2015deep}, IMDB-Wiki~\cite{rothe2018deep}, AwA 2~\cite{lampert2009learning}, and CUB~\cite{wah2011caltech} datasets have multiple binary classification tasks, while only one non-binary classification task. It is challenging to study the influence of complexity of classification tasks using these datasets and hence is not extendable to practical applications. CLEVR~\cite{johnson2017clevr} dataset provides with four different tasks with variable number of classes. However, each image contains multiple objects with different shape, color, and textures, allowing multiple labels for each task. Task suppression in multi-label, multi-task classification setting provides a very challenging experimental setting. 

\begin{figure*}[!t]
	\begin{center}
		\includegraphics[width=6.4in]{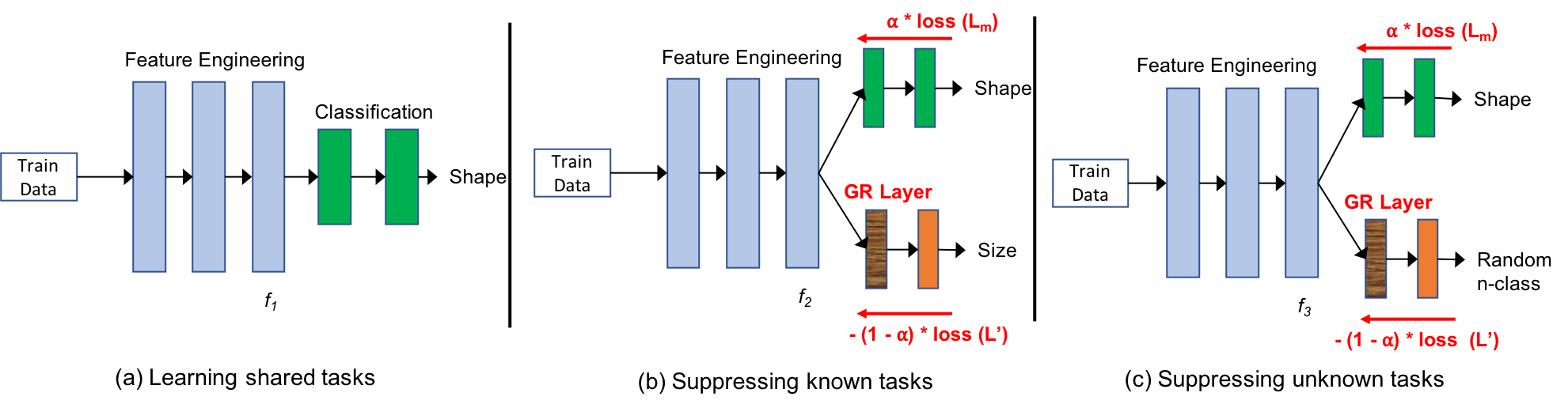}
	\end{center}
	\caption{(a) A deep learning model learning features suited for multiple tasks, more than the intended shape classification task, (b) Existing approaches suppress other known tasks, such as size classification by backpropagation of negative loss or gradient, (c) Proposed approach of suppressing all possible n-class classification task by using random class labels.}
	\label{fig:approach}
\end{figure*}

Inspired from the CLEVR dataset, we create a new \textit{PreserveTask} dataset, which is a multi-task dataset exclusively designed for the purpose of bench-marking models against preserving task privacy. The primary objective is to create easy-to-perform multi-task dataset, where the performance of the individual tasks is high. As shown in Figure~\ref{fig:dataset}, \textit{PreserveTask} dataset has five different classification tasks, as follows: (i) \textit{Shape Classification (5):} circle, triangle, diamond, pentagon, hexagon, (ii) \textit{Color Classification (7):} violent, indigo, blue, green, yellow, orange, red, (iii) \textit{Size Classification (3):} small, medium, large, (iv) \textit{Location Classification (4):} quadrant 1, quadrant 2, quadrant 3, quadrant 4, (v) \textit{Background Color Classification (3):} white, black, or colored.

These five tasks are chosen such that few tasks are highly correlated (size, shape), while few tasks are ideally independent of each other (size, color).
All the images are generated as $256 \times 256$ colored images. There are $5$ (shapes) * $7$ (color) * $3$ (size) * $4$ (location) * $3$ (background color) = $1260$ variations, with $50$ images for training and $10$ images for testing for each variation, generating a total of $63,000$ training and $12,600$ images. This ensures that there is a perfect class balance across all tasks. It is to be noted that the task of suppression of unknown shared task is a fairly open research problem. Hence, in order to set the benchmark of different frameworks, an easy, straight-forward \textit{PreserveTask} dataset is created as a conscious decision without having much noise, such as in DeepFashion~\cite{liu2016deepfashion} dataset. As the problem area matures, further extensions of this dataset could be generated and more real world natural objects could be added.
\section{Proposed Approach}

To understand the current scenario of model overleanring, consider any deep learning model as shown in Figure~\ref{fig:approach} (a). Assume a deep learning model, say VGG19, is trained for predicting the shape of objects in images. Ideally, the features $f_1$ obtained from the model should be good for object shape prediction. However, it is observed that $f_1$ has highly entangled features with other tasks such as size, color, and location. This enables us to train prediction classifiers for other tasks on top of $f_1$ without the need for the original data. In literature, few technique variants exist to suppress the model from learning a few attributes or tasks~\cite{narayanaswamy2017learning, madras2018learning, edwards2015censoring}. As shown in Figure~\ref{fig:approach} (b), if the model has to be suppressed from learning the size of the objects, a negative loss or negative gradient is applied to enable features $f_2$ to not carry any information about the size of the object while retaining all the information about the shape of the object. This comes with an assumption that the information and class labels about the tasks to be suppressed are available during training time for the entire training data. 

In our proposed framework, we overcome this assumption and do not expect the suppression task information to be available during model training time. Additionally, we provide a model agnostic approach of suppressing task overlearning so that the framework could be directly applied to any deep learning model.
Let $x \in X$ be the input data and $y_x^{(1)} \in Y^{(1)}$ to $y_x^{(n)} \in Y^{(n)}$ be the $n$ different tasks that could be performed on the image. We learn a model, $g(f(x)): X\xrightarrow{}Y^{(i)}$, where $f(.): X \xrightarrow{} Z^{(i)} $, be the feature representation for the given task, $i$. Ideally, while only $g(.): Z^{(i)}\xrightarrow{} Y^{(i)}$ should be possible, we observe that $g(.): Z^{(i)}\xrightarrow{} Y^{(j)}$ for $i \neq j$ provides high classification accuracy in most cases. To overcome this challenge, we generate random n-class labels to simulate any possible n-class classification task, that need to be suppressed. These random labels generated for an unknown task are provided in the gradient reversal (GR) branch~\cite{ganin2014unsupervised} in order to suppress any other n-class classification, as shown in Figure~\ref{fig:approach} (c). Multiple gradient reversal branches could be built for varying values of $n$ to suppress all possible other classification tasks. The DL model is trained by a custom loss function as follows,

\begin{multline}
 \mathcal{L}^{(i)}(\theta_p, \theta_s) = \sum_{x \in X} \left[ \lambda . \mathcal{L}_{\theta_p}(y_x^{(i)}, g(f(x))) \right. \\ \left. - (1-\lambda) \mathcal{L}_{\theta_s}(rand(\mathbb{R}^{Y^{(j)}}), g(f(x))) \right]
\end{multline}

where $L_{\theta_p}$ is the loss of the model branch trained for the task, $i$, to be preserved. $L_{\theta_s}$ is the loss obtained from the other branch which needs to be maximized (task suppression). $y_x^{(i)}$ is the actual ground truth label for the sample $x$ for task $i$. $rand(\mathbb{R}^{Y^{(j)}}$ generated a random class label in the space of $Y^{(j)}$, where $j \neq i$ and $unique(Y^{(i)}) \neq unique(Y^{(j)})$. $\lambda$ is the regularization parameter controlling the weight given for the minimization and maximization losses and is a hyperparameter chosen manually based on the amount of sharing between the tasks. $L_{\theta_p}$ and $L_{\theta_s}$  could be any choice of the popular loss functions, depending on number of classes, classification/ regression tasks, and multi-label classification. Thus, it can be observed that the proposed framework is both DL model agnostic and loss function agnostic.

\subsection{Trust Score}

\begin{figure*}[!t]
	\begin{center}
		\includegraphics[width=6.4in]{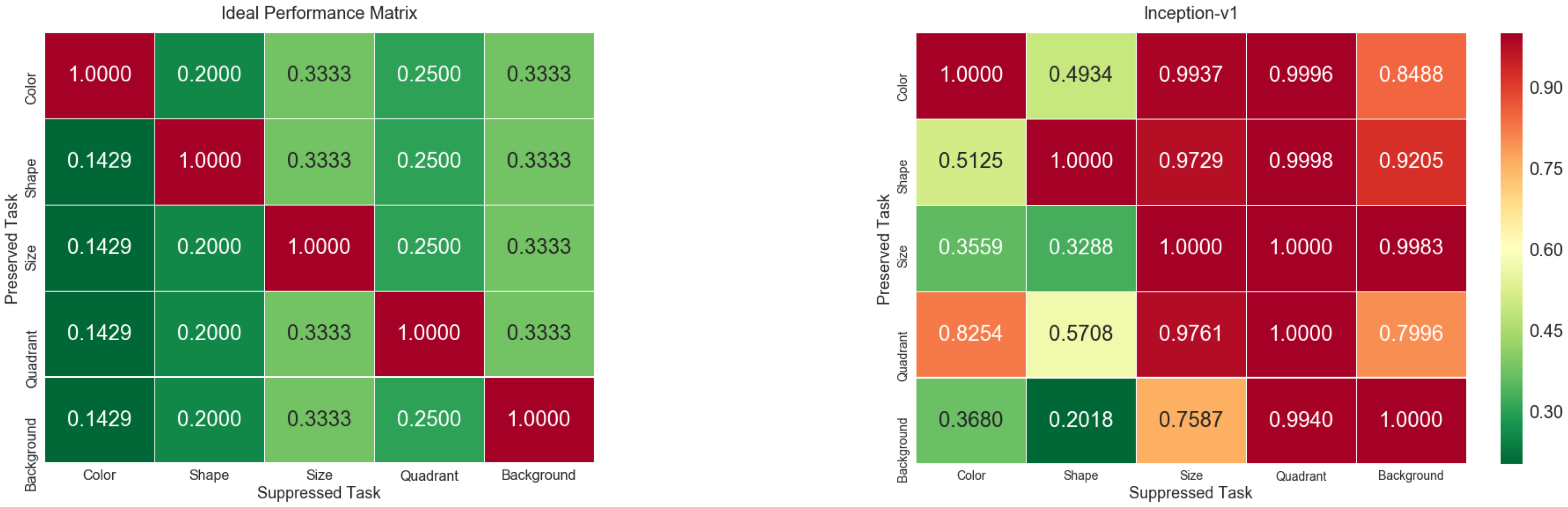}
	\end{center}
	\caption{(Left) The accuracy matrix demonstrating the behavior of an ideal trusted DL model. The leading diagonal shows perfect classification while the rest of the values are random classification. (Right) The accuracy matrix detailing the shared task performance of Inception-v1 on the \textit{PreserveTask} dataset.}
	\label{fig:results1}
\end{figure*}

\textit{PreserveTask} will be used as the benchmark dataset against which the trust score of any trained DL model could be extracted. The trained DL model is evaluated against different tasks in the \textit{PreserveTask} and the entire confusion matrix of performance accuracy is obtained ($5 \times 5$ corresponding to the five tasks). The behavior of an ideal DL model, would provide $100\%$ accuracy on the leading diagonal i.e., the tasks it was trained for, while providing, random classification accuracies for other tasks. The confusion matrix for such an ideal DL model is shown in Figure~\ref{fig:results1}. For example in the first row, the DL model was trained to learn and predict the color of the object. Hence, color prediction performance should be $1$ (denoting, $100\%$ accuracy), while other tasks should provide random $1/n$ accuracy, where $n$ is the number of classes. 

Let the ideal performance matrix be denoted as $M$ and the obtained performance matrix for a given trained DL model be $T$. By intuition, the matrix $T$ that does not deviate much from the ideal matrix $M$ should have a higher trust score. The trust score is mathematically computed as follows,
\begin{equation}
    \text{Trust Score} = 1 - \frac{\sum_i(|M - T| \cdot W )}{\sum_i W}
\end{equation}

where, $W = (n_t - 1) \times \mathcal{I}_{n_t} \cdot \mathbb{1}_{n_t}$ provides the weight corresponding to each task pair, $I$ is an identity matrix and $\mathbb{1}_{n_t}$ is a ones matrix, each of dimensionality $n_t\times n_t$, where $n_t$ is the number of tasks including the preserved and suppression tasks. In \textit{PreserveTask} dataset, $n_t = 5$ resulting in a $W$ matrix with leading diagonal elements to be $4$ while the rest of the elements to be $1$.  
Since for each preserving task, there are four suppressing tasks, the deviation of the preserving task from the ideal matrix is scaled by a factor of four to normalize the computation. Also, $|M-T|$ represents the absolute difference between $M$ and $T$ matrices, and $\sum_i(.)$ is the sum of all elements in the matrix.

Note that if the diagonal elements perform poorly, the concern is on the performance of the model. On the contrary, if the non-diagonal elements has a higher performance, the concern is on the trust of a model from a privacy preservation perspective. The proposed metric implements this notion to compute the trustworthiness of a trained DL model. The trust score is bounded between [0,1]. By empirical analysis, we observe that a trust score above $0.9$ is highly desirable, a trust score between $0.8$ and $0.9$ is practically acceptable, and any score below $0.8$ is considered poor. The trust score of the ideal matrix is $1$, while the trust score of a $\mathbb{1}_5$ (all task classification performance is $100\%$) is $0.6259$.
To understand the sensitivity of the proposed metric, let us assume that in the ideal matrix, any one non-diagonal element is changed to $1$ which results in a trust score of $0.98125$. Thus, any reduction of ($1$ - $0.98125$) = $0.0175$ in the trust score corresponds to one additional task being overlearnt by the classifier.

\section{Experimental Results}

In this section, we show the experimental results and perform analysis of the proposed framework. Initially, we measure the trustworthiness of the existing models. We then experimentally demonstrate suppression of different tasks in various experimental settings. All the experiments are performed using the \textit{PreserveTask} dataset. For additional results and detailed comparison with other techniques, please refer to the appendix.

\begin{figure*}[!t]

	\begin{center}
		\includegraphics[width=6.4in]{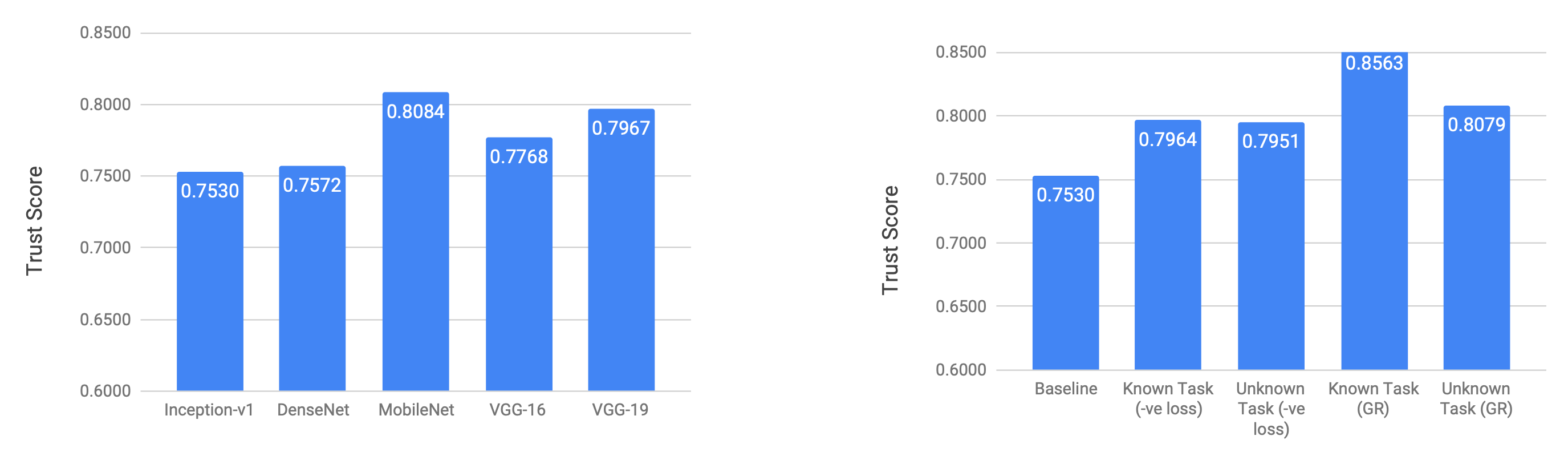}
	\end{center}
	\caption{(Left) Trust scores obtained for various DL models. It can be observed that, of the five models, the Inception-v1 and MobileNet has the least and highest trust score, respectively. (Right) Trust scores obtained after various suppression techniques for Inception-v1. It can be observed that using random labels for unknown tasks, we could improve the trustworthiness.}
	\label{fig:trust_1}
\end{figure*}

\begin{figure*}[!t]
	\begin{center}
		\includegraphics[width=6.5in]{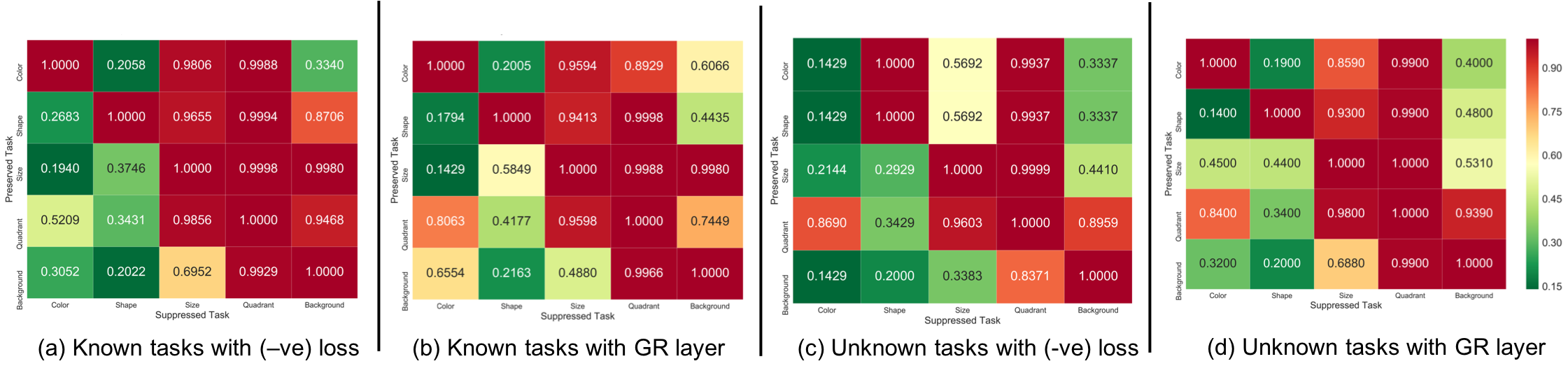}
	\end{center}
	\caption{The performance matrix obtained after suppressing the known tasks in (a), (b) and the unknown tasks in (c), (d). Comparative results between a baseline negative loss function and the proposed GR layer based suppression is also shown. All results are computed for the Inception-v1 model.}
	\label{fig:all_confusion}
\end{figure*}

\subsection{How Trustworthy are Existing Models?}
Consider a popular deep learning model, Inception-v1~\cite{szegedy2016rethinking} consisting of 22 computational layers. The model was trained from scratch using the \textit{PreserveTask} for the task of shape classification, providing $99.98\%$. In order to study, if this deep learning model learnt additional visual attributes, as well, the last flatten layer's output ($4096\times 1$) were extracted. Four different two-hidden layer neural network classifiers ($512$, $100$) were trained~\footnote{with default scikit-learn parameters} using the extracted features to predict size, color, location, and background color of the objects. The prediction accuracies were $97.29\%$, $51.25\%$, $99.98\%$, $92.05\%$, respectively for the four tasks. It can be observed that the performance of size, location, and background prediction are really high proving that the features obtained from Inception v1 model has features corresponding to these tasks as well. Also, it can be observed that the color prediction performance is very low, as shape and color prediction are inherently independent tasks. The similar experiment is repeated for training the Inception v1 model on one task and using the learnt feature to predict the performance of other tasks, and the results are shown in Figure~\ref{fig:results1}. Ideally, only the diagonal elements of this confusion matrix should have higher accuracies (red in color) while the rest of the prediction should have lower accuracies (green in color). Accordingly, the trust score of the trained Inception-v1 model (proposed in section 4.1) was found to be $0.7530$, which is very poor. 

In order to further demonstrate that this additional intelligence is not a property of just Inception-v1 model, similar experiments are performed using four other popular deep learning models: VGG16, VGG19, MobileNet, and DenseNet. The trust scores of all the DL models are shown in Figure~\ref{fig:trust_1} (a). It can be observed that out of these five models, Inception-v1 and DenseNet has the lowest trust score while MobileNet has the highest trust score. While one could argue that the Inception-v1 model learns highly generic features supporting multi-task and transfer learning, from a privacy preservation perspective, the model is found to have a poor trust score. This leads to the open question, ``\textit{Do models always needs to be additionally intelligent, and if not, how to suppress them?}"

\subsection{How to Suppress Known Tasks?}

In this section, we perform experiments to suppress the tasks that are known apriori during training, that is, the ground truth labels of the suppression task is available. For simplicity, in demonstrating the experimental results, we assume that one task is to be preserved and one task is to be suppressed, using the Inception-v1 model. This experimental setting is similar to the approach explained in Figure~\ref{fig:approach} (b). The gradient reversal (GR) layer unlearns the suppressed task, while learning the preserved task. In order to compare the performance of GR, we also use a customized negative loss function which minimizes the loss obtained for the preserved task while maximizing the loss obtained for the suppressed task, weighted by a constant factor. The features eventually extracted from the flatten layer has to show similar performance on the preserved task while reduced performance on the suppressed task. 

Figure~\ref{fig:all_confusion} (a) and (b) demonstrates the results obtained for Inception-v1 using negative loss function and the proposed GR layer. While the leading diagonal elements showed the same performance, in comparison with Figure~\ref{fig:results1}, it can be observed that prediction results of the suppressed tasks reduced in most of the cases. For example, while preserving the object shape prediction, suppressing the background color prediction performance dropped from $92.05\%$ to $44.35\%$. This indicates that the extracted features no longer contain information about the background color of the image. The corresponding trust scores are shown in Figure~\ref{fig:trust_1} (b). It can be observed that suppressing known tasks using GR layer improves the trust of the baseline model from $0.7530$ to $0.8563$.

\begin{figure}[!t]

	\begin{center}
		\includegraphics[width=3in]{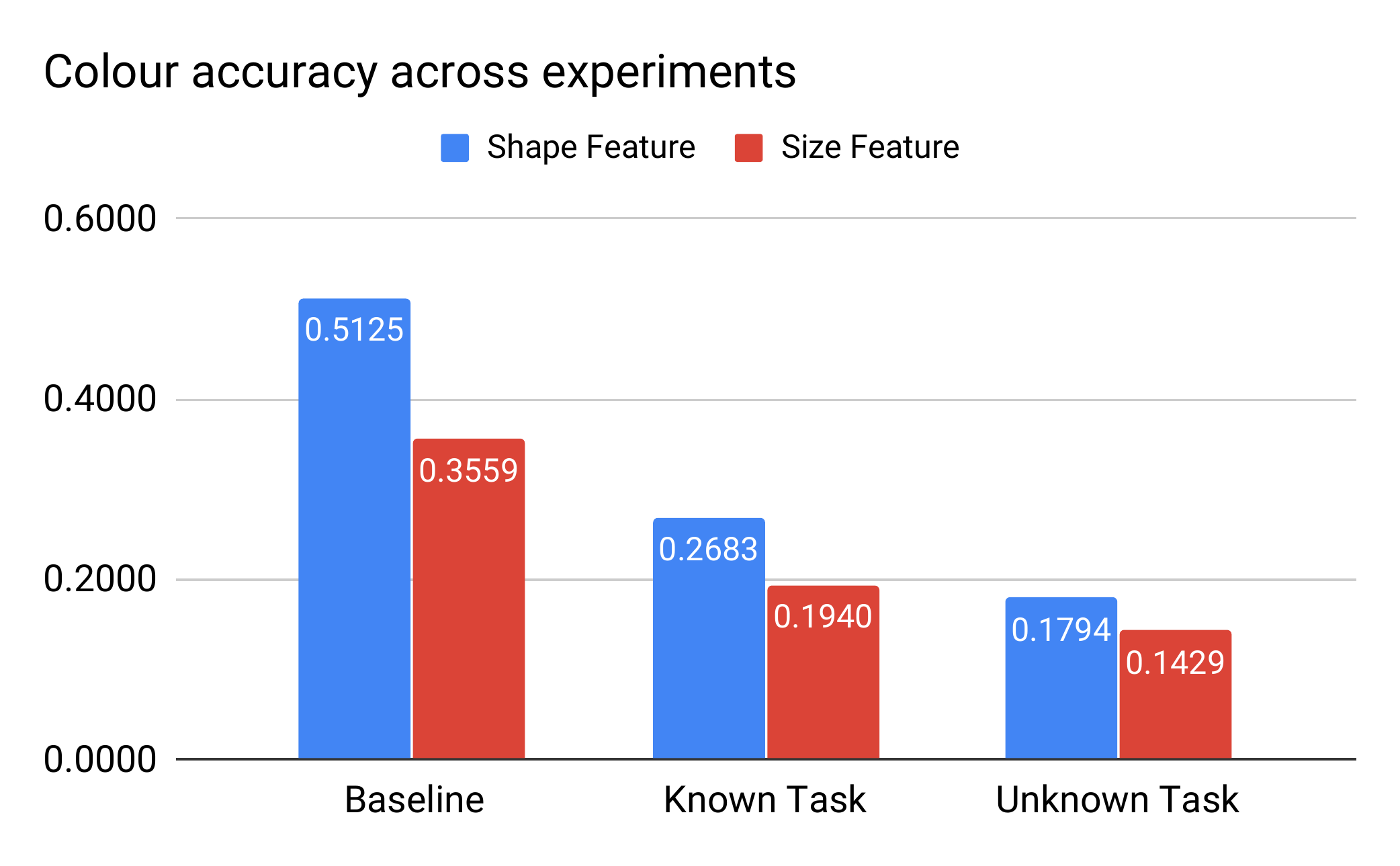}
	\end{center}
	\caption{Comparison of color prediction performance with and without using the different task suppression mechanisms. It can be observed that using random labels reduces the performance of color prediction irrespective of whether the preserved task was shape or size prediction.}
	\label{fig:results4}
\end{figure}

\subsection{How to Suppress Unknown Tasks?}

The results obtained in the previous section made the assumption that the ground truth labels of the suppression task have to be available while training the Inception-v1 model. In an attempt to break that assumption, the experimental setting discussed in Figure~\ref{fig:approach} (c) is performed. Instead of the actual ground truth labels of a particular task, randomly generated n-class labels are used during every mini-batch. Thus, for the same mini-batch training in the next epoch, a different set of random class labels are generated to be maximized. This ensures that the model does not memorize a single suppression task, but, learns to suppress all possible n-class classification tasks.

\begin{figure*}[!t]
	\begin{center}
		\includegraphics[width=6.4in]{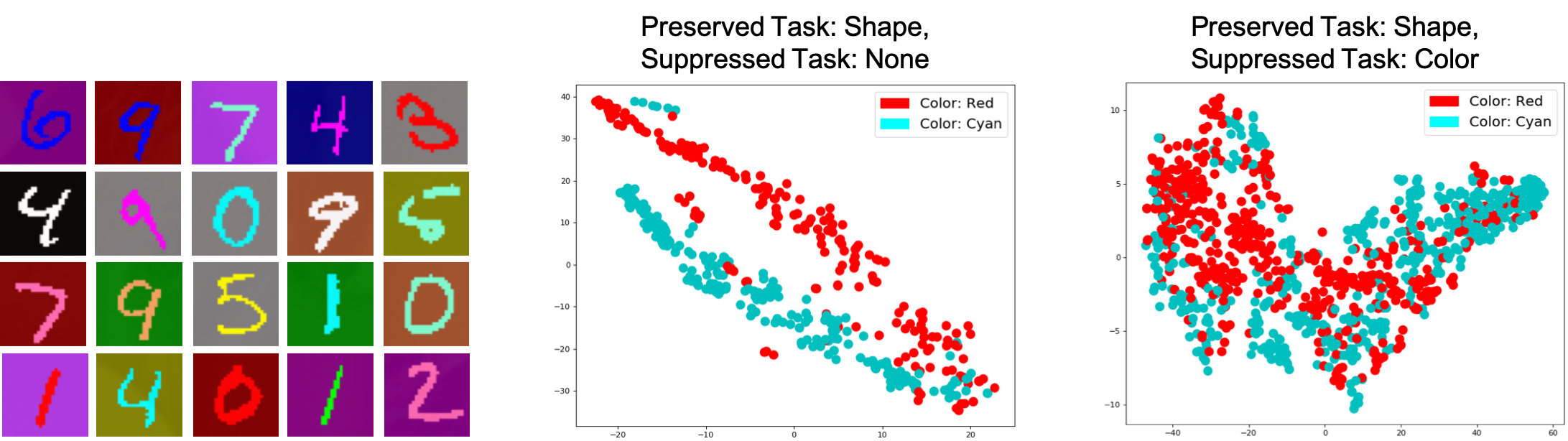}
	\end{center}
	\caption{(Left) Sample images from the colored MNIST dataset. (Right) TSNE plot of the feature distribution of 392 images (class 0, foreground color: red and cyan) before and after suppressing the color prediction task.}
	\label{fig:color_mnist}
\end{figure*}

Figure~\ref{fig:all_confusion} (c) and (d) demonstrates the results obtained by using random class labels. In comparison with Figure~\ref{fig:results1}, it can be observed that using random class performs well in certain settings. For example, while trying to preserve the shape features and suppressing the prediction capacity of background color, the original model's prediction performance of $92.05\%$ reduced to $87.06\%$ by using the actual labels of background color, while further reduced to $33.37\%$ while using random 3-class labels. It is further highlighted in Figure~\ref{fig:results4} where color prediction is chosen as the task to be suppressed, while shape and size are independently being preserved.
It can be observed that the proposed framework of using random labels, reduces the performance of color prediction from $51.25\%$ to $26.83\%$ when using actual labels and $17.94\%$ when using random labels, when shape prediction was the preserved task. A similar performance reduction from $35.59\%$ to $14.29\%$ is observed when size prediction was the preserved task. 

We conclude that using random labels for task suppression produces a comparable trust score to using known labels while producing surely better results than the baseline trust score of a DL model.

\section{Case Study on Challenging Practical Datasets}
\textbf{Colored MNIST Dataset:}
We introduced two additional tasks of foreground and background color prediction tasks into the MNIST dataset. As shown in Figure~\ref{fig:color_mnist}, colored MNIST images are created by randomly assigning one of the $10$ possible foreground colors and one of the different $10$ possible background colors. Similar assignment is performed in both training and test dataset, to maintain the standard experimental protocol. MobileNet model was trained from scratch to obtained a baseline trust score of $0.756$.  After using our framework for task suppression with random labels and gradient reversal based training on the suppression branch, we observed that the MobileNet model’s trust scores increased to $0.824$. In Figure~\ref{fig:color_mnist} (middle), the TSNE plot shows that when the model is learnt only for shapes, the features for `red' and `cyan` colored images are still separable. However, after suppressing the color prediction task using the proposed framework, the features `red' and `cyan` colored images are scattered and no longer separable, as shown in Figure~\ref{fig:color_mnist} (right). 

\textbf{Diversity in Faces (DiF) Dataset:}
In DiF dataset~\cite{dif_arxiv19}, we considered the tasks of gender (two class) and pose (three class) classification. The aim is learn (preserve) only one of these while suppressing the other. Since, the dataset was highly skewed for different classes, we considered a subset of $39296$ images with equal class balance\footnote{Please refer to the appendix for the exact data distribution and the detailed performance matrix obtained}. We trained Inception-v1 model on this dataset from scratch and obtained a trust score of $0.7497$. Using our framework for task suppression with GR layer and known class labels, the trust score of the model increased to $0.8606$. Additionally, with random unknown class labels, we observed that the model's trust scores increased to $0.9069$.

\textbf{IMDB-Wiki Dataset:}
In IMDB-Wiki dataset~\cite{rothe2018deep}, we considered the tasks of gender (two class) and age (ten class) classification. The cropped face images of the Wiki dataset are used to train the DenseNet model (the second least trusted model according to our trust scores). The trained model provided a baseline trust score of $0.7846$. After using our framework for task suppression and known class labels, the trust score of DenseNet model increased to $0.7883$. Also, with random unknown class labels, we observed that the model's trust scores increased to $0.7860$.

Thus, our framework for measuring and improving a DL model's trust has lots of practical applications. A face recognition system or a face image based gender recognition system can now be deployed with an additional trust on the model's intelligence level.

\section{Conclusion and Future Research}

We showcased a model-agnostic framework for measuring and improving the trustworthiness of a model from a privacy preservation perspective. The proposed framework did not assume the need for the suppression task labels during train time, while, similar performance could be obtained by training using random classification boundaries. A novel simulated benchmark dataset called \textit{PreserveTask} was created to methodically evaluate and analyze a DL model's capability in suppressing shared task learning. This dataset opens up further research opportunities in this important and practically necessary research domain. Experimentally, it was shown that popular DL models such as VGG16, VGG19, Inception-v1, DenseNet, and MobileNet show poor trust scores and tend to be more intelligent than they were trained for. Also, we show a practical case study of our proposed approach in face attribute classification using: (i) Diversity in Faces (DiF) and (ii) IMDB-Wiki datasets.
We would like to extend this work by studying the effect of multi-label classification tasks during suppression.

{\small
\bibliographystyle{ieee_fullname}
\bibliography{bib}
}

\clearpage
\section{Appendix}

This supplementary material contains all the detailed hyper-parameters used by different models that we trained, to aid in reproducing the results that we showed in the research paper. Additionally, we provide more detailed analysis and visualizations of the results, that could not be included in the paper due to space constraints.

\subsection{Baseline Deep Learning Models}
Five different baseline deep learning models were used in the experiments: Inception-v1, VGG16, VGG19, DenseNet, and MobileNet. The different parameters and the training process used in these experiments are shown below:
\begin{itemize}
    \item The data is z-normalized to have a zero mean and unit standard deviation, before being provided to the models for training.
    \item The standard architectures of Inception-v1, VGG16, VGG19, DenseNet, and MobileNet are borrowed from the default implementations in the Keras library.
    \item The deep learning models were trained with \textit{categorical cross-entropy} and \textit{Adam} optimizer with parameters as learning rate = $0.0001$ and amsgrad set as $False$.
\end{itemize}

\subsection{Classifier Models}
For all the experiments, a two hidden layer neural network is used as a classifier. This is to maintain consistency of the same classifier across all the experiments. 
\begin{itemize}
    \item The architecture is Dense (512) $\rightarrow$ Dropout (0.5) $\rightarrow$ Dense (100) $\rightarrow$ Dropout (0.3) $\rightarrow$ Dense (num\_of\_classes)
    \item Each of the Dense layer has a $ReLU$ activation function.
    \item \textit{categorical cross-entropy} is used as the loss function with \textit{Adam} as the optimizer, having parameter values as  learning rate = $0.0001$ and amsgrad set as $False$.
    \item $20\%$ of the data is used as validation data and the model is trained for $100$ epochs with early stopping. 
    \item Batch size of $32$  was used to make the computation faster and the experiments were run using $1\times K80$ GPU. 
    
\end{itemize}

\section{Experimental Results and Analysis}

In this section, we are including additional analysis, visualizations, and charts of the results presented in the main paper. In order to aid better comparison, we include the charts and results presented in the main paper also here, so that the supplementary could be read in an independent manner.

\subsection{How Trustworthy are Existing Models?}

\begin{figure}[h]
	\begin{center}
		\includegraphics[width=3.4in]{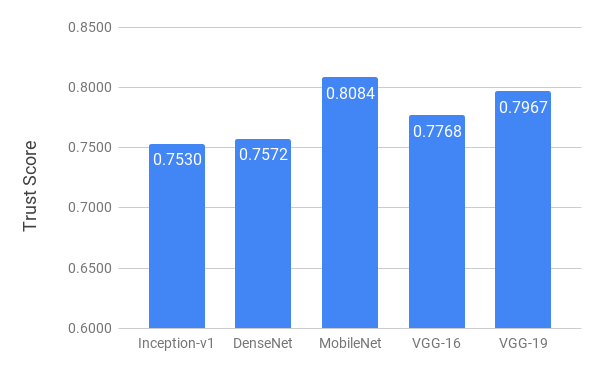}
	\end{center}
	\caption{Trust scores obtained for various DL models. It can be observed that, of the five models, the Inception-v1 and DenseNet has the least trust score while MobileNet has the highest.}
	\label{fig:trust_1}
\end{figure}

\begin{figure}[h]
	\begin{center}
		\includegraphics[width=3.2in]{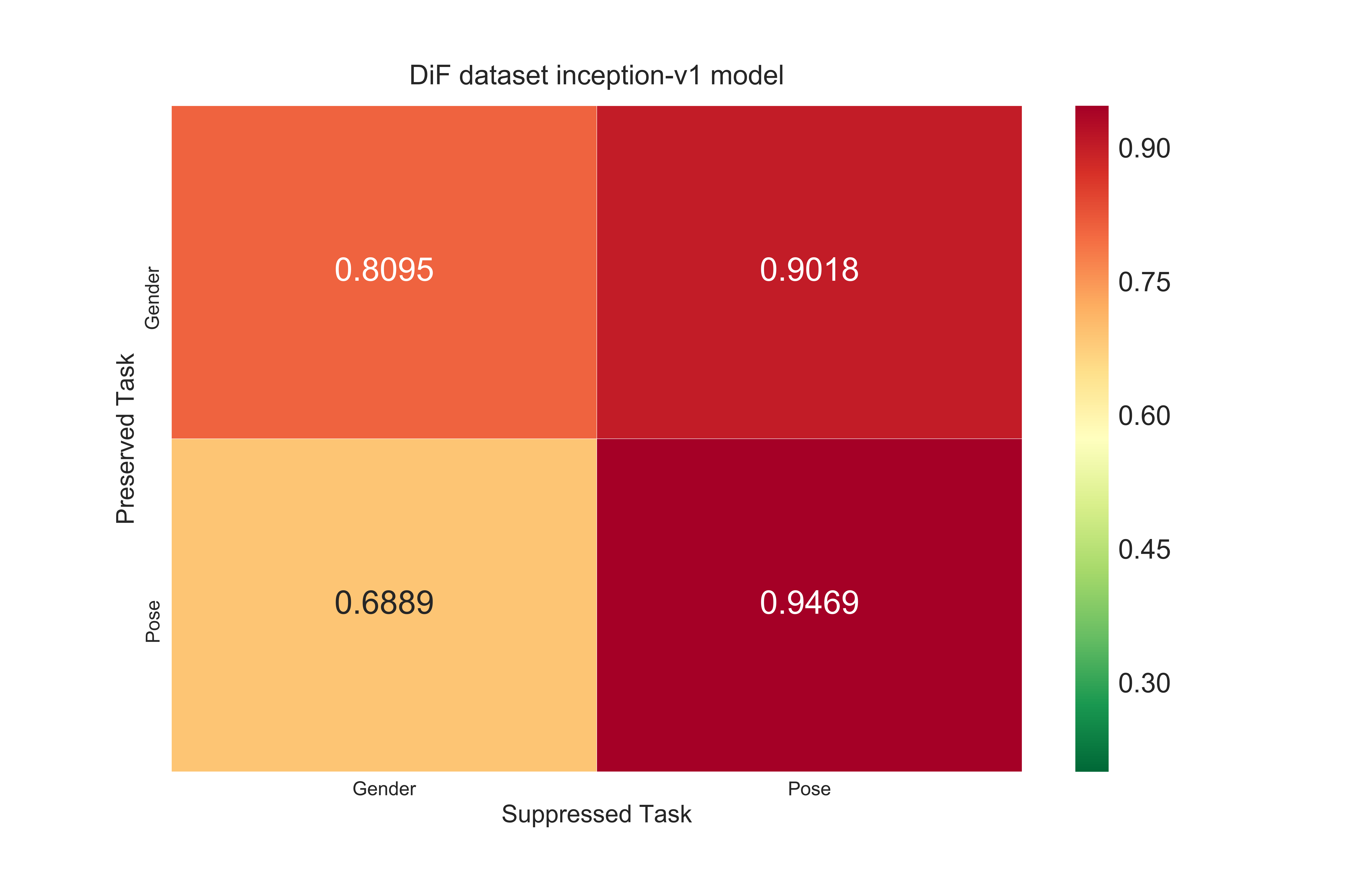}
	\end{center}
	\caption{The performance matrix heat-map detailing the shared task performance of Inception-v1 model on the \textit{PreserveTask} dataset.}
	\label{fig:results11}
\end{figure}

\begin{figure}[h]
	\begin{center}
		\includegraphics[width=3.2in]{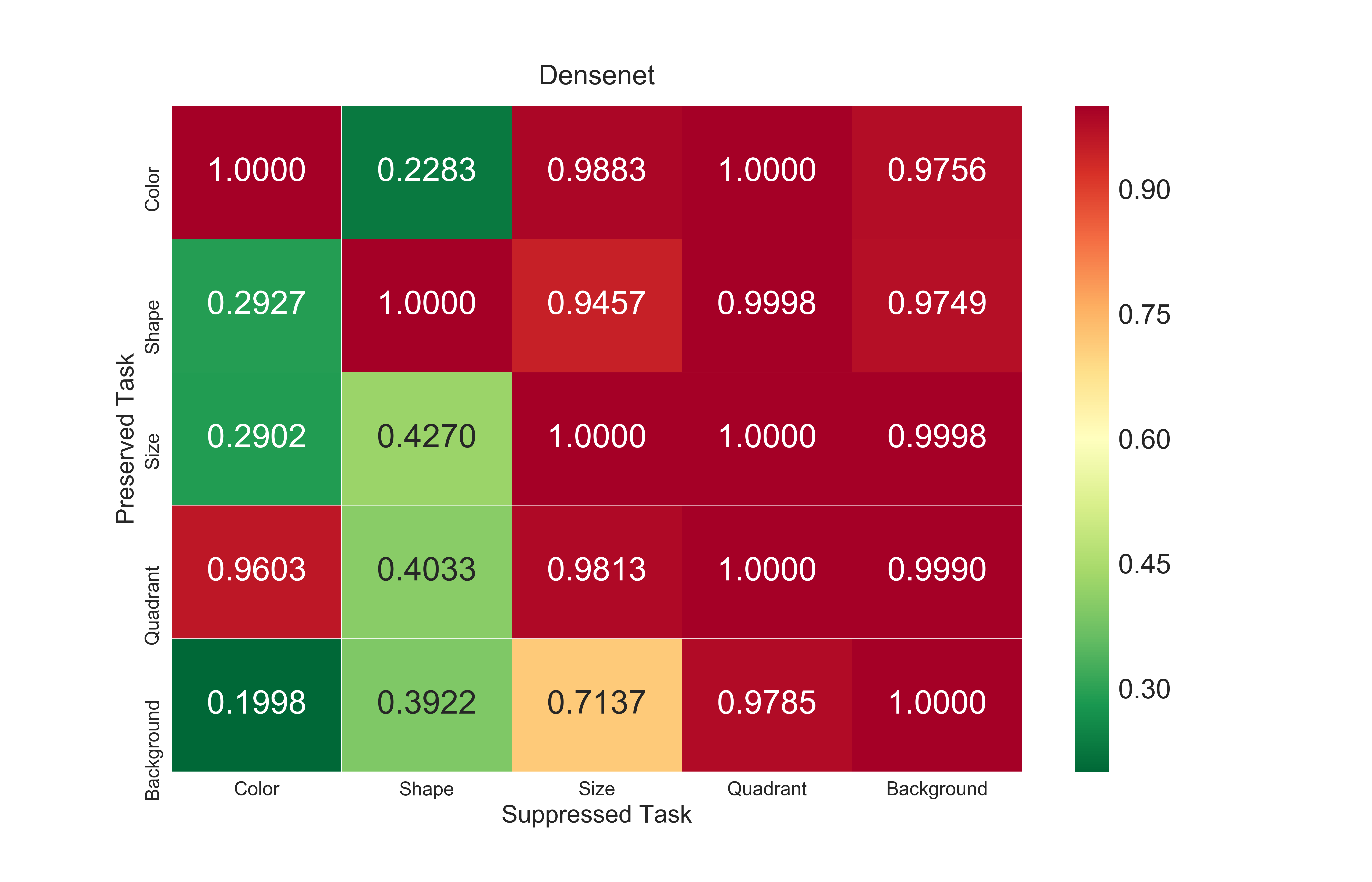}
	\end{center}
	\caption{The performance matrix heat-map detailing the shared task performance of DenseNet model on the \textit{PreserveTask} dataset.}
	\label{fig:results12}
\end{figure}

\begin{figure}[h]
	\begin{center}
		\includegraphics[width=3.2in]{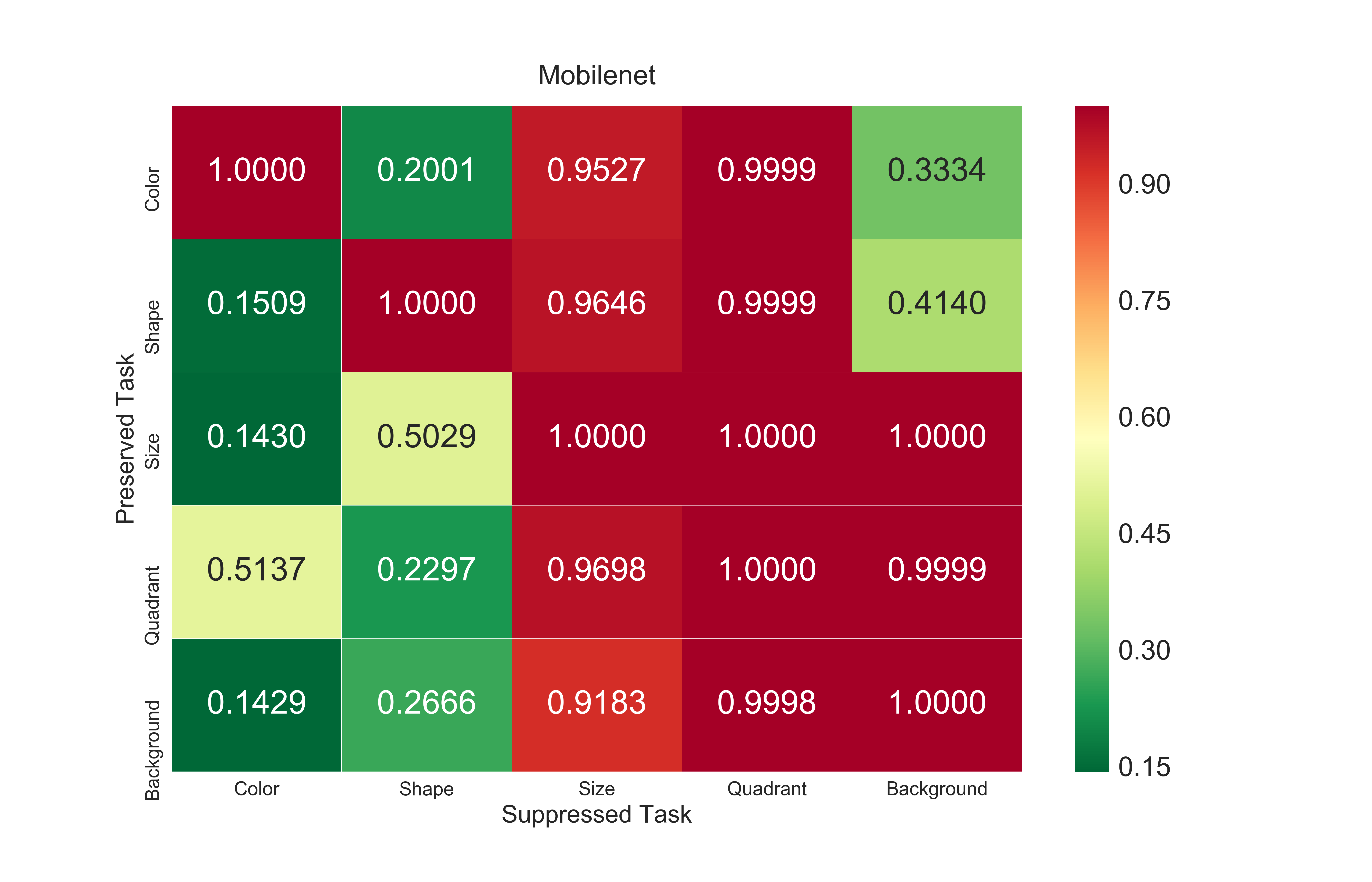}
	\end{center}
	\caption{The performance matrix heat-map detailing the shared task performance of MobileNet model on the \textit{PreserveTask} dataset.}
	\label{fig:results13}
\end{figure}

\begin{figure}[h]
	\begin{center}
		\includegraphics[width=3.2in]{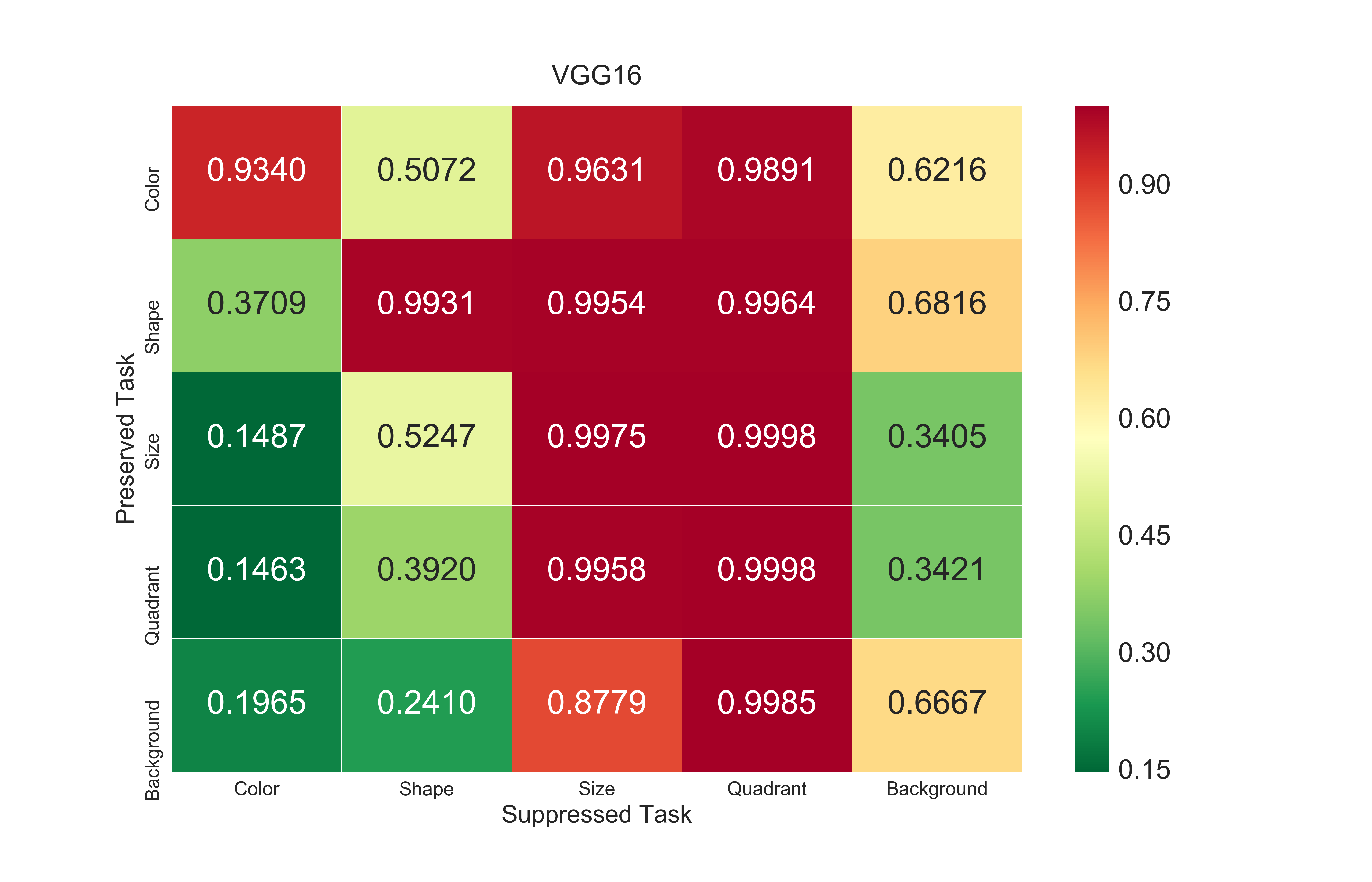}
	\end{center}
	\caption{The performance matrix heat-map detailing the shared task performance of VGG-16 model on the \textit{PreserveTask} dataset.}
	\label{fig:results14}
\end{figure}

\begin{figure}[h]
	\begin{center}
		\includegraphics[width=3.2in]{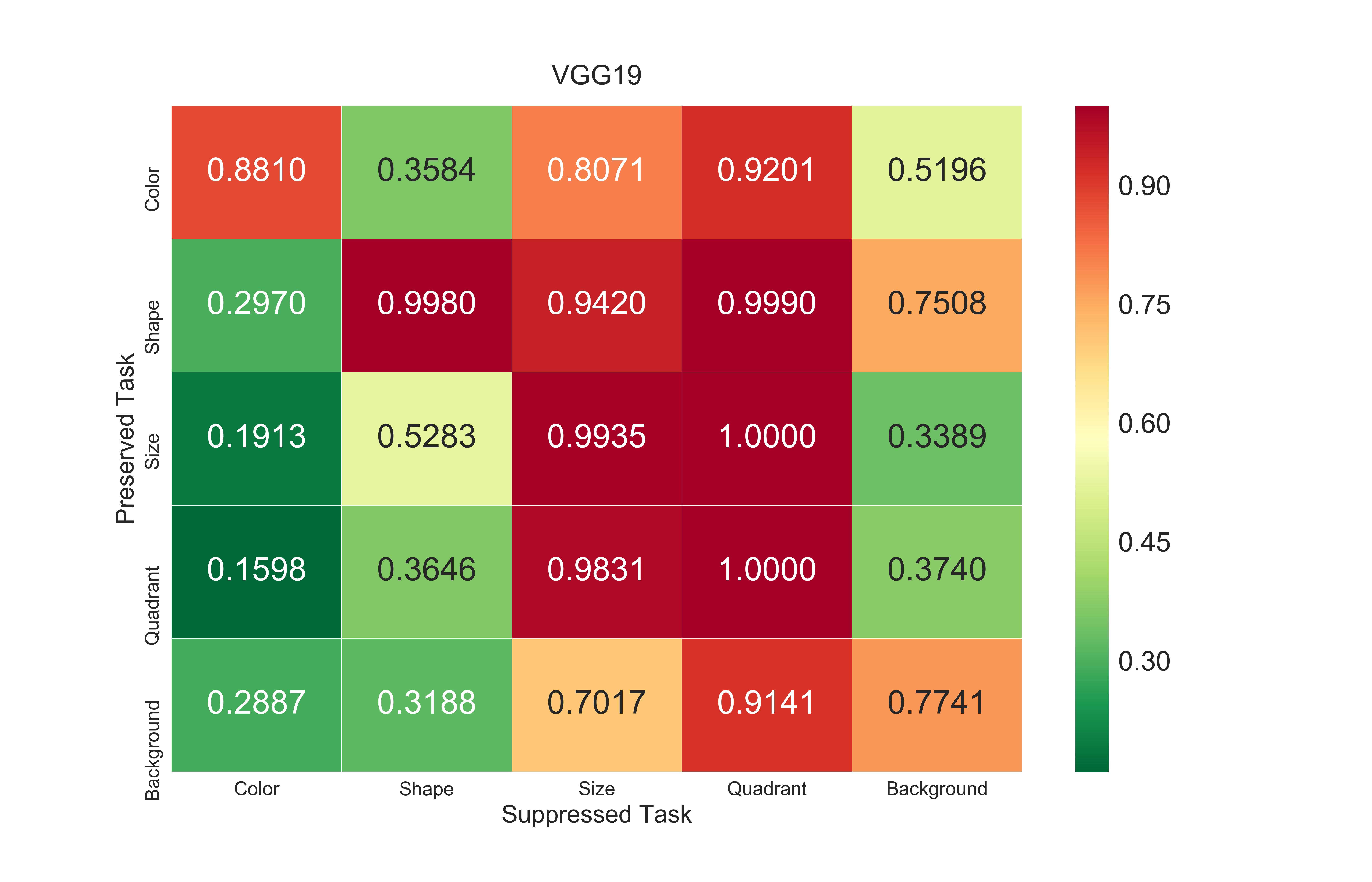}
	\end{center}
	\caption{The performance matrix heat-map detailing the shared task performance of VGG-19 model on the \textit{PreserveTask} dataset.}
	\label{fig:results15}
\end{figure}

\newpage

\subsection{How to Suppress Tasks?}

\begin{figure}[h]

	\begin{center}
		\includegraphics[width=3.4in]{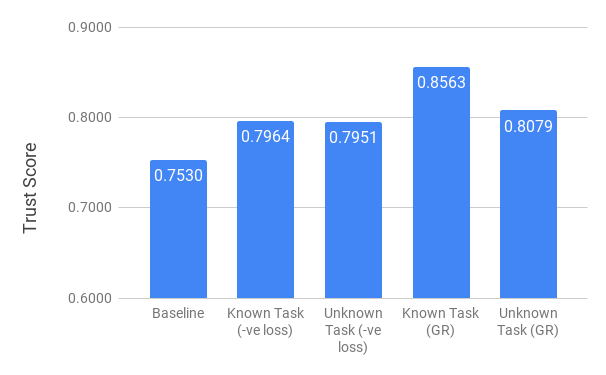}
	\end{center}
	\caption{Trust scores obtained after various suppression techniques. It can be observed that even using random labels for unknown tasks, we could improve the trustworthiness of the Inception-v1 model on the \textit{PreserveTask} dataset.}
	\label{fig:trust_2}
\end{figure}

\begin{figure}[h]
	\begin{center}
		\includegraphics[width=3.2in]{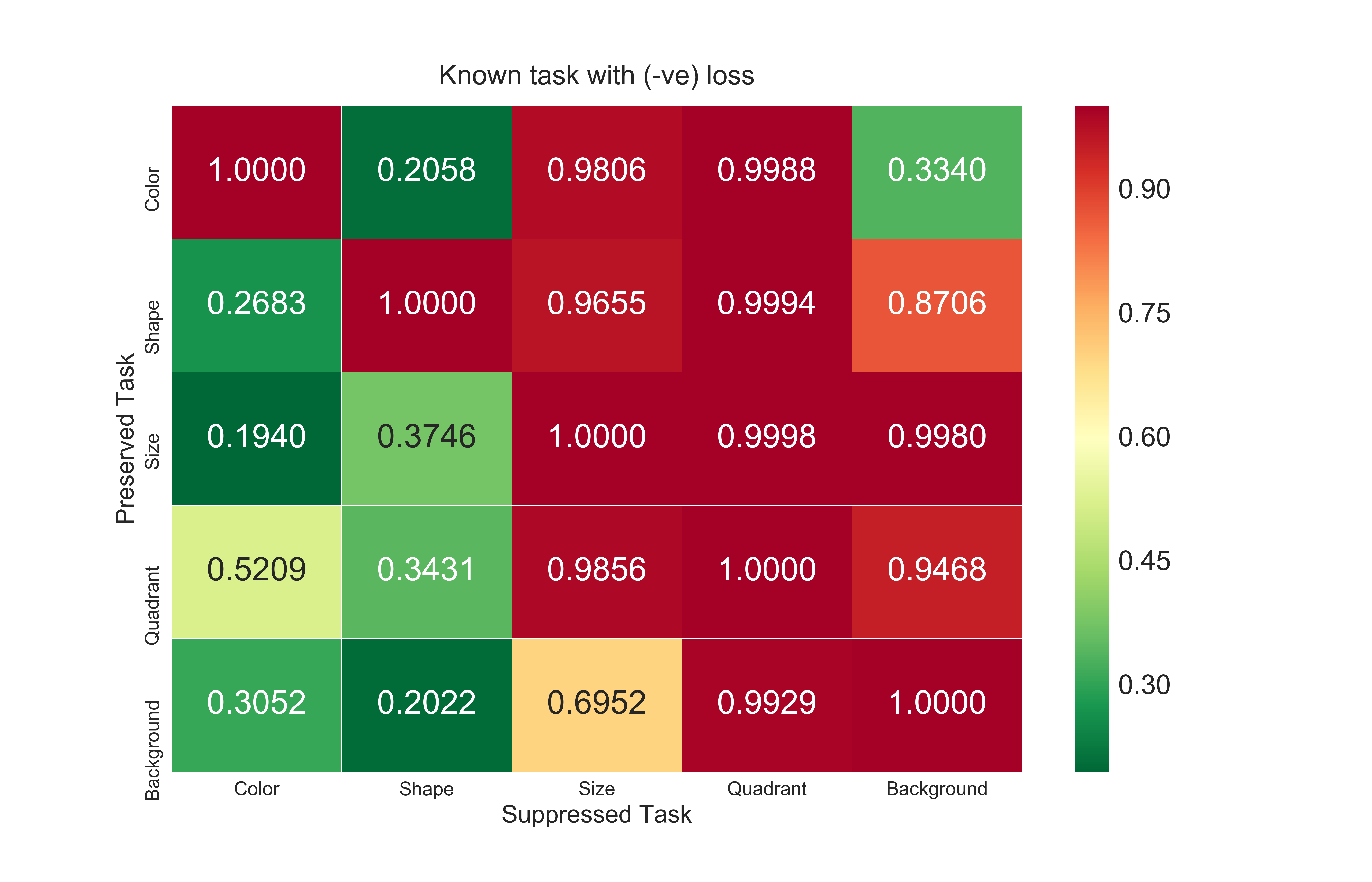}
	\end{center}
	\caption{The performance matrix heat-map, after suppressing a known task using negative loss, detailing the shared task performance of Inception-v1 model on the \textit{PreserveTask} dataset.}
	\label{fig:results21}
\end{figure}

\begin{figure}[h]
	\begin{center}
		\includegraphics[width=3.2in]{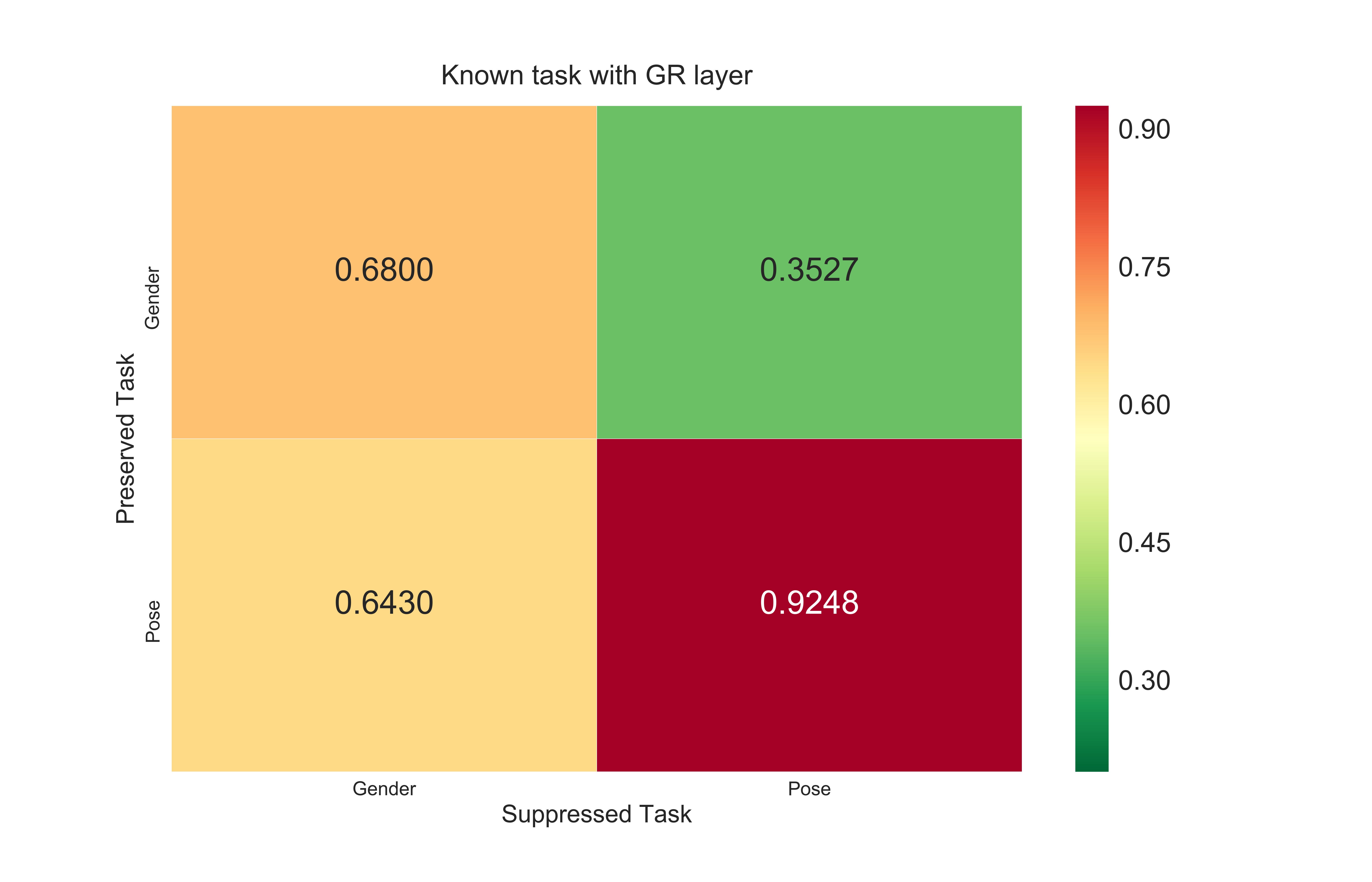}
	\end{center}
	\caption{The performance matrix heat-map, after suppressing a known task using GR layer, detailing the shared task performance of Inception-v1 model on the \textit{PreserveTask} dataset.}
	\label{fig:results22}
\end{figure}

\begin{figure}[h]
	\begin{center}
		\includegraphics[width=3.2in]{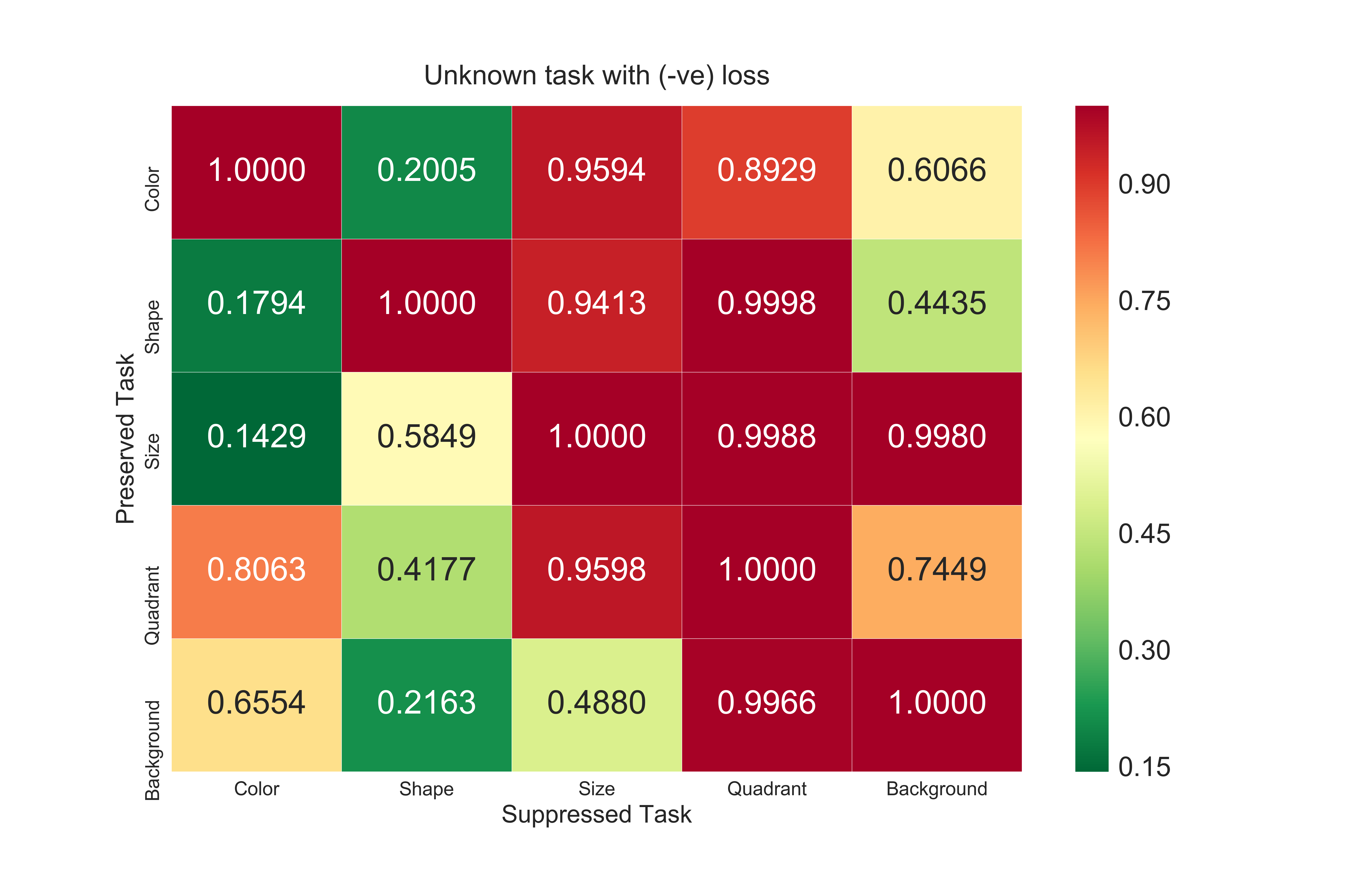}
	\end{center}
	\caption{The performance matrix heat-map, after suppressing a unknown task using negative loss, detailing the shared task performance of Inception-v1 model on the \textit{PreserveTask} dataset.}
	\label{fig:results23}
\end{figure}

\begin{figure}[h]
	\begin{center}
		\includegraphics[width=3.2in]{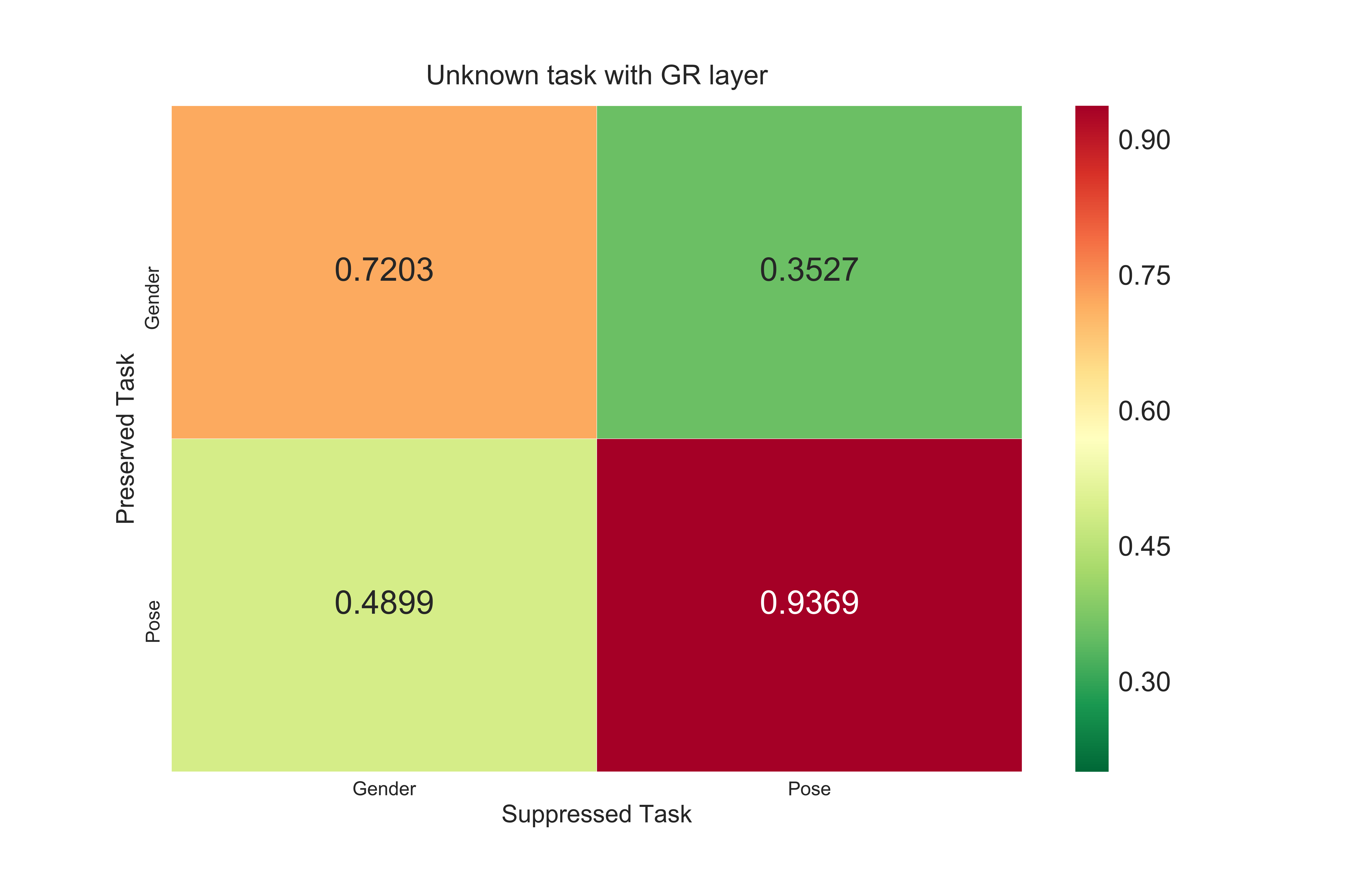}
	\end{center}
	\caption{The performance matrix heat-map, after suppressing a unknown task using GR layer, detailing the shared task performance of Inception-v1 model on the \textit{PreserveTask} dataset.}
	\label{fig:results24}
\end{figure}

\newpage

\section{Case Study: Face Attribute Preservation}

\begin{figure}[h]
	\begin{center}
		\includegraphics[width=3.4in]{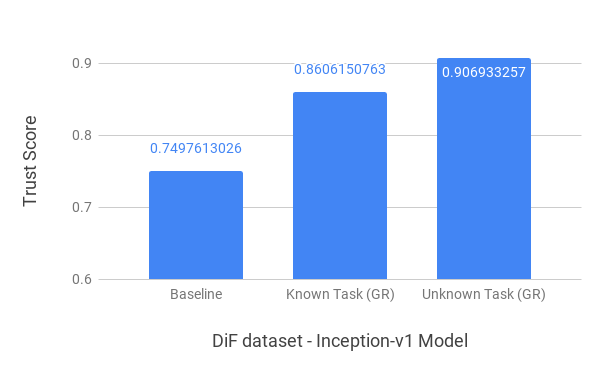}
	\end{center}
	\caption{Trust scores obtained in the Diversity in Faces (DiF) dataset after various suppression techniques. It can be observed that even using random labels for unknown tasks, we could improve the trustworthiness of the Inception-v1 model.}
	\label{fig:trust_3}
\end{figure}

\begin{figure}[h]
	\begin{center}
		\includegraphics[width=3.2in]{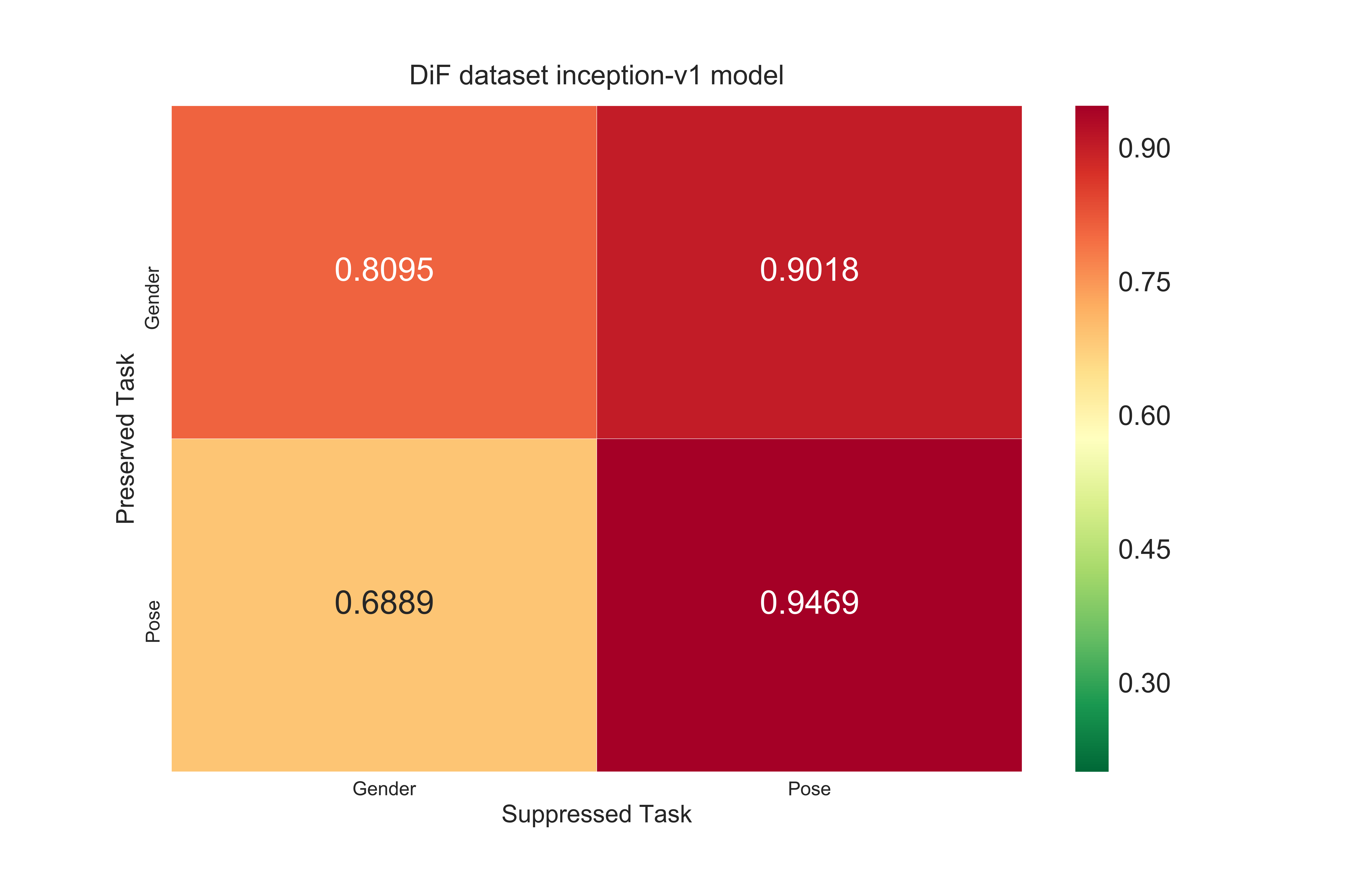}
	\end{center}
	\caption{The performance matrix heat-map detailing the shared task performance of Inception-v1 model on the Diversity in Faces (DiF) dataset.}
	\label{fig:results1657}
\end{figure}

\begin{figure}[h]
	\begin{center}
		\includegraphics[width=3.2in]{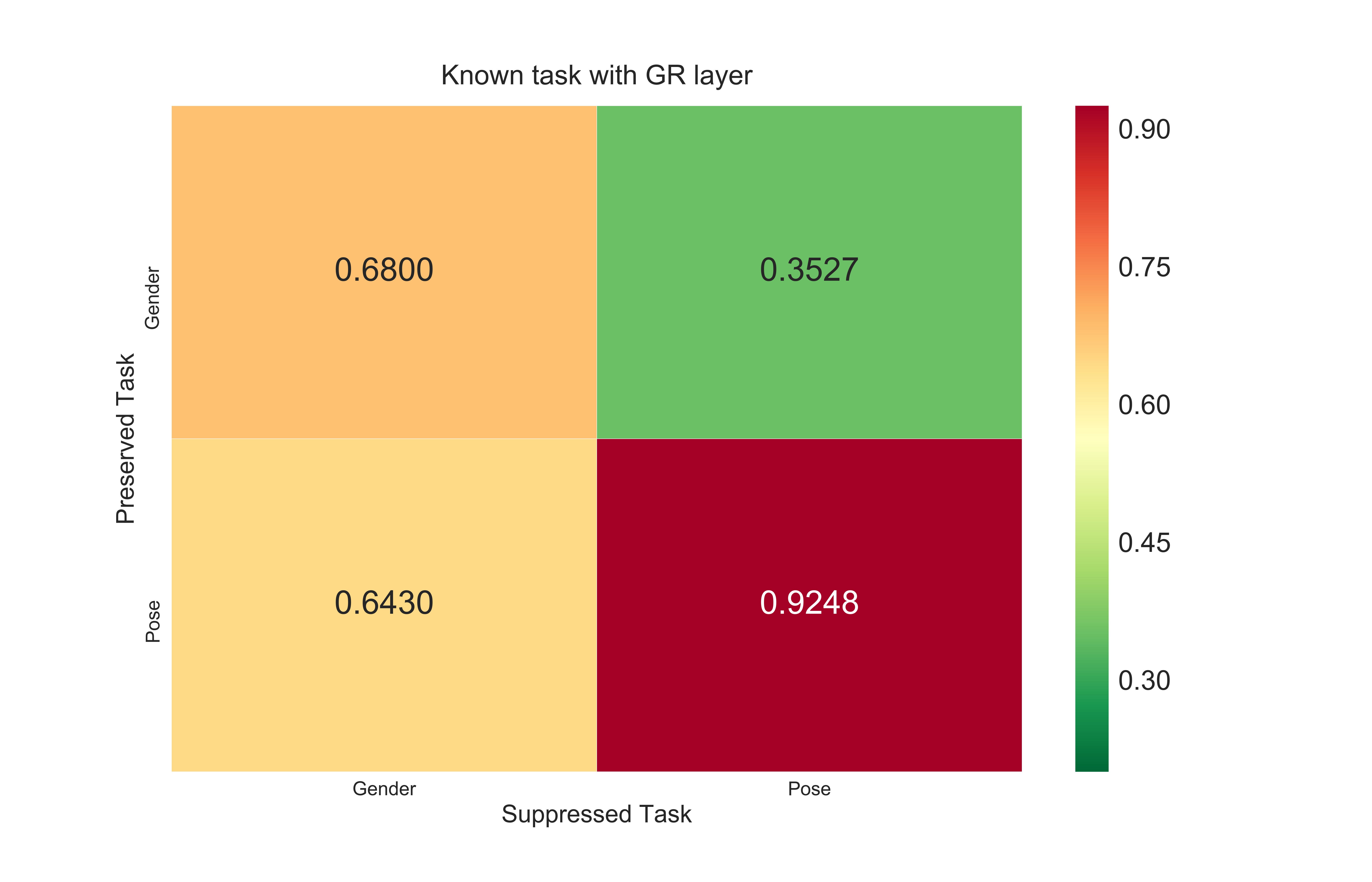}
	\end{center}
	\caption{The performance matrix heat-map obtained after suppressing the known tasks, detailing the shared task performance of Inception-v1 model on the Diversity in Faces (DiF) dataset.}
	\label{fig:results2}
\end{figure}

\begin{figure}[h]
	\begin{center}
		\includegraphics[width=3.2in]{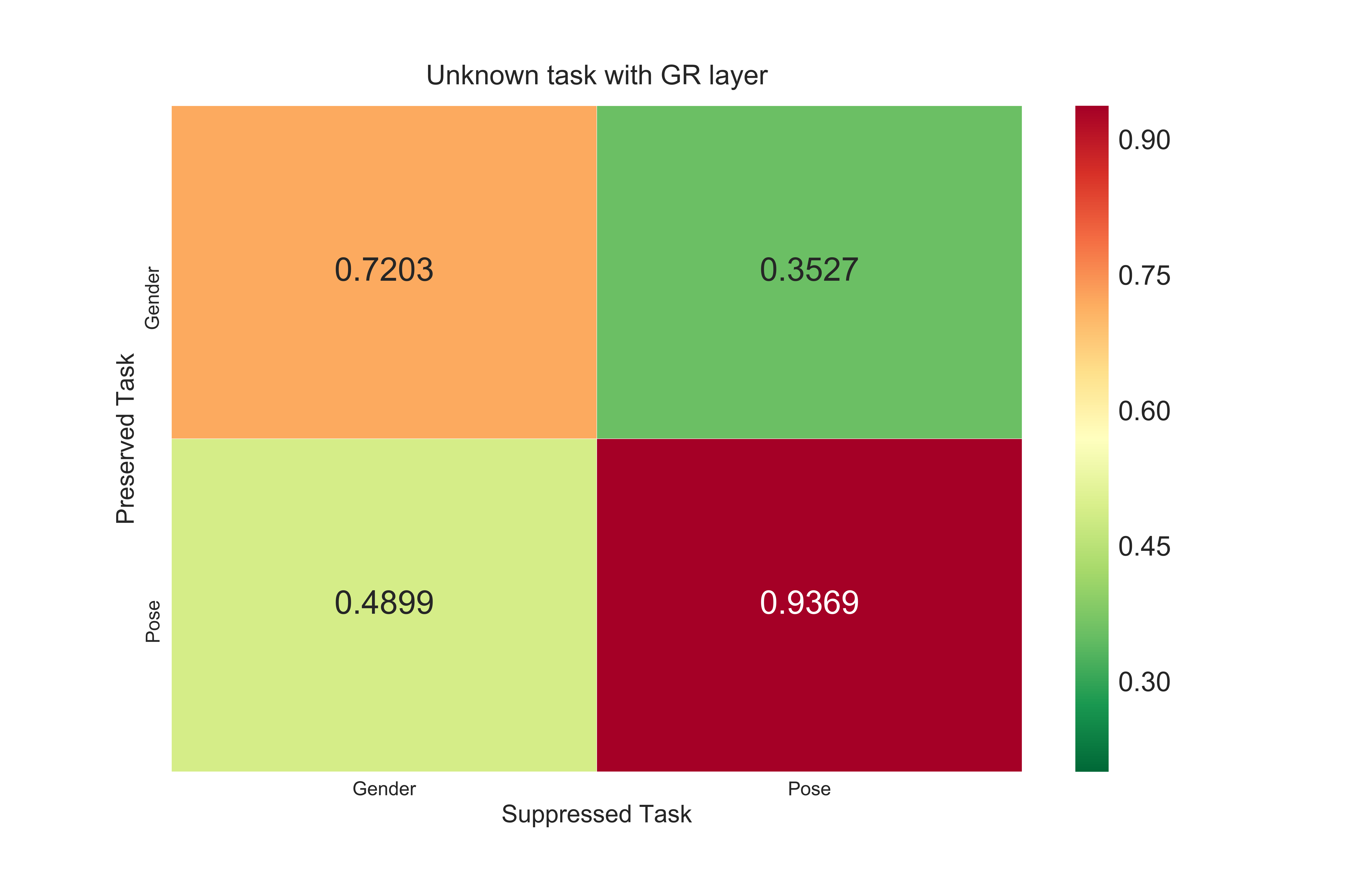}
	\end{center}
	\caption{The performance matrix heat-map obtained after suppressing the unknown tasks, detailing the shared task performance of Inception-v1 model on the Diversity in Faces (DiF) dataset.}
	\label{fig:results3}
\end{figure}

\begin{figure}[h]

	\begin{center}
		\includegraphics[width=3.4in]{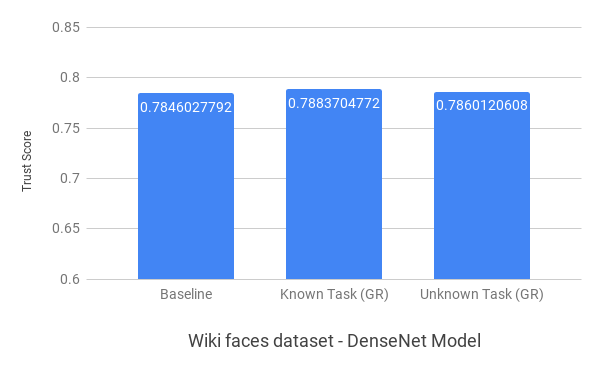}
	\end{center}
	\caption{Trust scores obtained in the WIKI face dataset after various suppression techniques. It can be observed that even using random labels for unknown tasks, we could improve the trustworthiness of the Inception-v1 model.}
	\label{fig:trust_4}
\end{figure}

\begin{figure}[h]
	\begin{center}
		\includegraphics[width=3.2in]{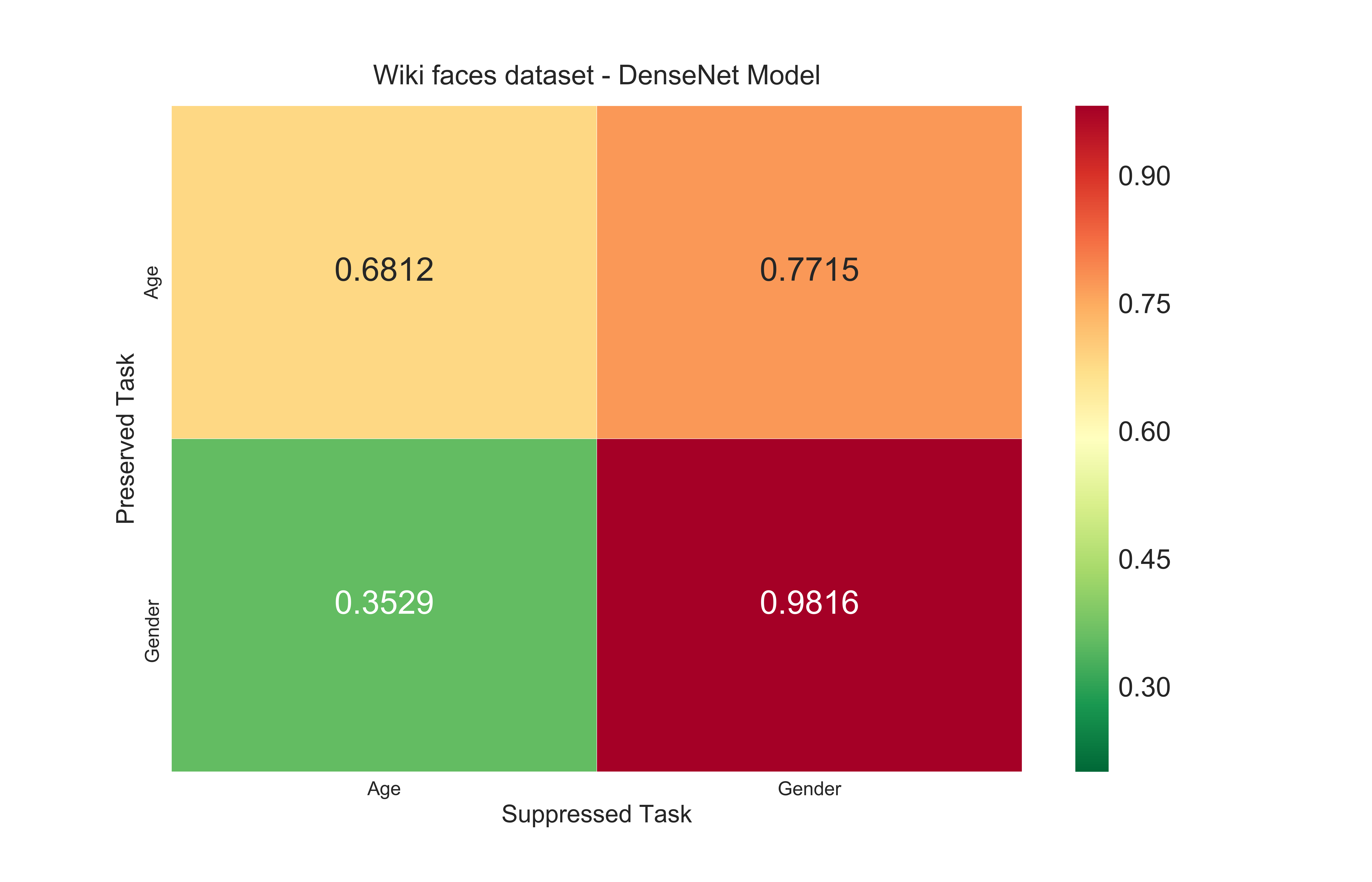}
	\end{center}
	\caption{The performance matrix heat-map detailing the shared task performance of DenseNet model on the Wiki face dataset.}
	\label{fig:results18798}
\end{figure}

\begin{figure}[h]
	\begin{center}
		\includegraphics[width=3.2in]{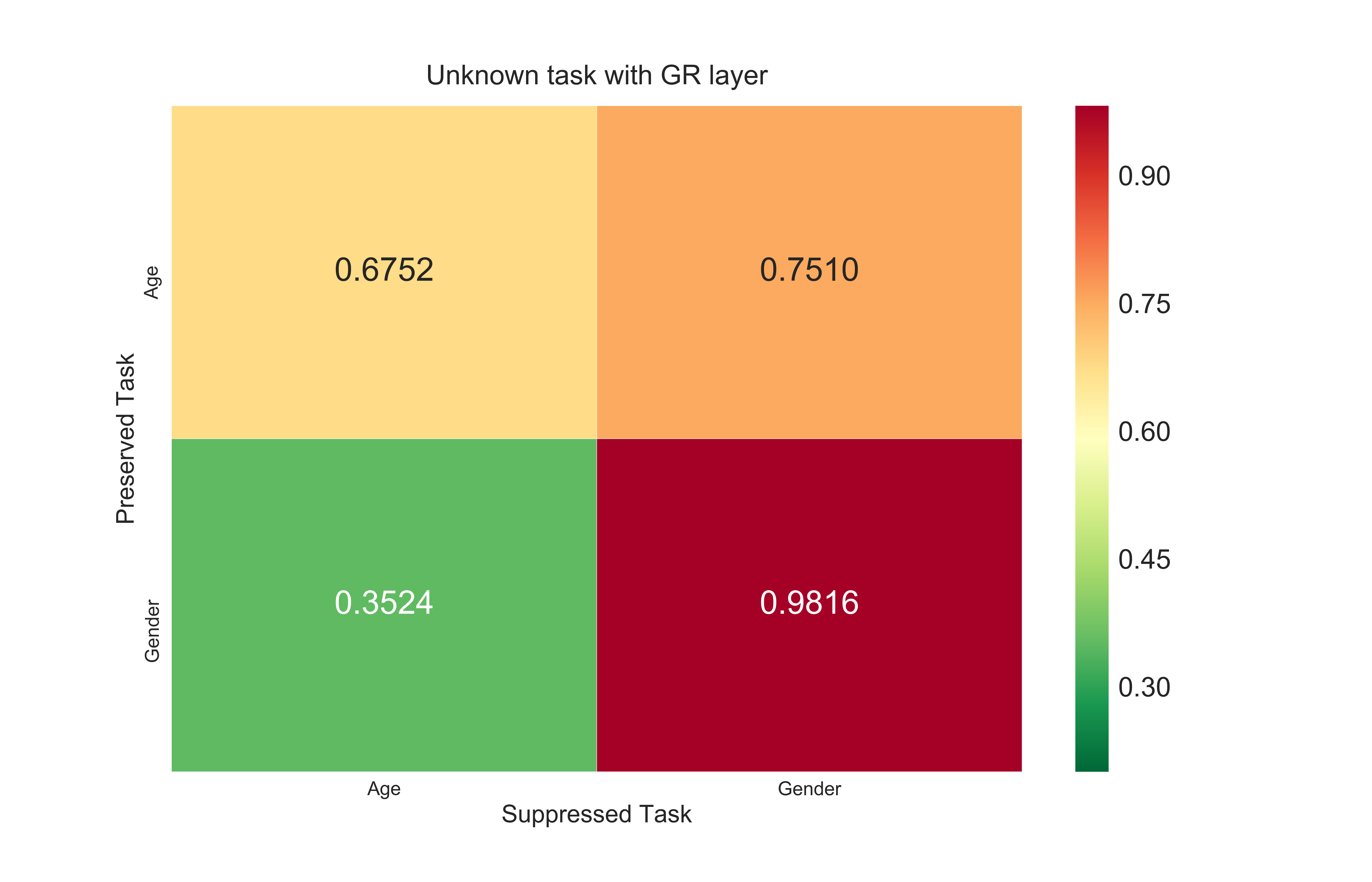}
	\end{center}
	\caption{The performance matrix heat-map obtained after suppressing the known tasks, detailing the shared task performance of DenseNet model on the Wiki face dataset.}
	\label{fig:results2}
\end{figure}

\begin{figure}[h]
	\begin{center}
		\includegraphics[width=3.2in]{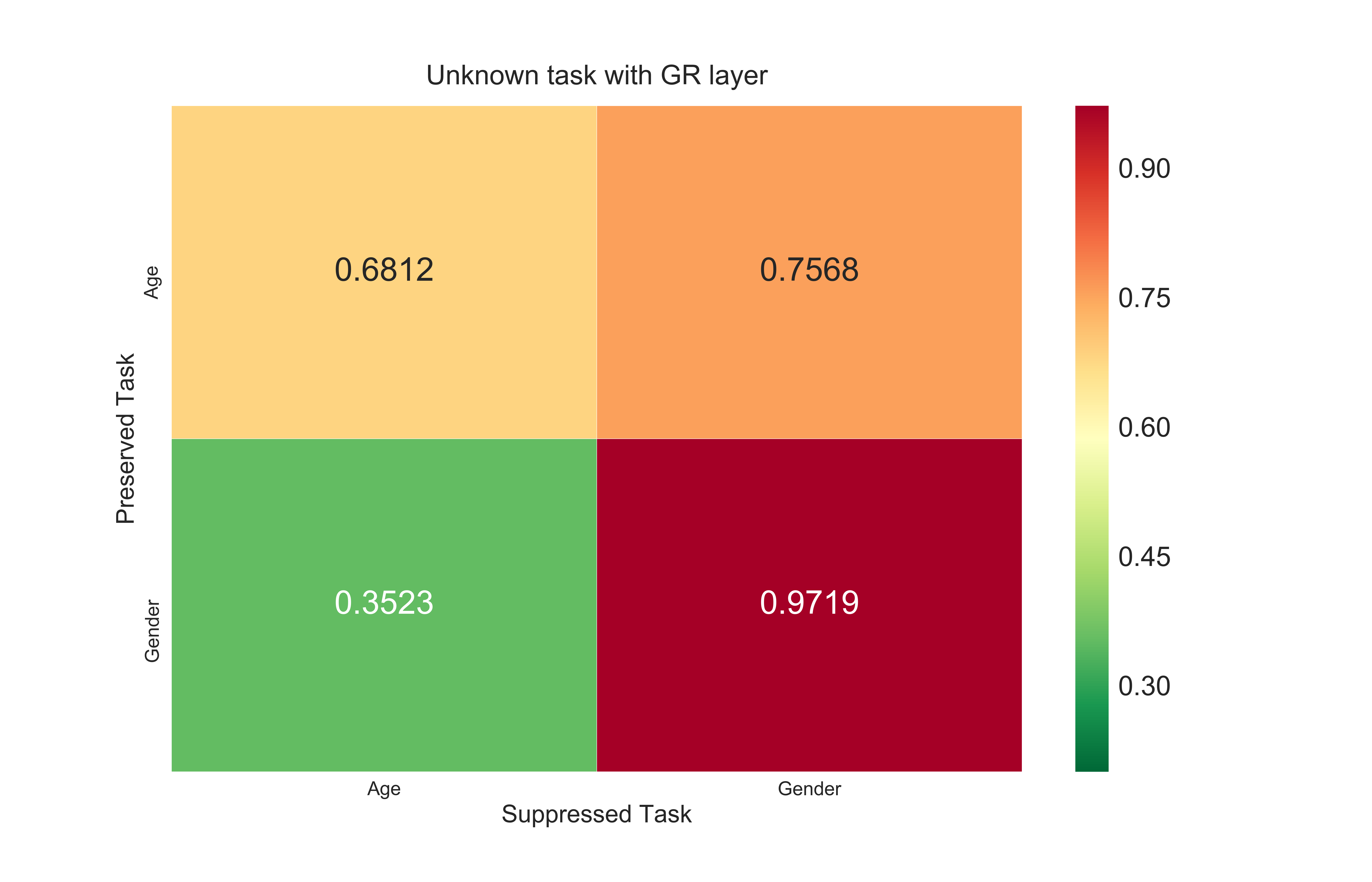}
	\end{center}
	\caption{The performance matrix heat-map obtained after suppressing the unknown tasks, detailing the shared task performance of DenseNet model on the Wiki face dataset.}
	\label{fig:results3}
\end{figure}

\end{document}